\def\year{2022}\relax
\documentclass[letterpaper]{article} 
\usepackage{aaai22}  
\usepackage{times}  
\usepackage{helvet}  
\usepackage{courier}  
\usepackage[hyphens]{url}  
\usepackage{graphicx} 
\usepackage{natbib}  
\usepackage{caption} 
\DeclareCaptionStyle{ruled}{labelfont=normalfont,labelsep=colon,strut=off} 
\usepackage{algorithm}
\usepackage{algorithmic}

\usepackage{newfloat}
\usepackage{listings}
\floatstyle{ruled}
\newfloat{listing}{tb}{lst}{}
\floatname{listing}{Listing}
\usepackage{xcolor}
\usepackage{amsmath,amssymb,amsfonts}
\usepackage{bbm}
\usepackage{graphicx}

\usepackage{booktabs}   
\usepackage{multirow}
\usepackage{diagbox}
\usepackage{xcolor}
\usepackage{csquotes}

\title{Graphical Models with Attention for Context-Specific Independence and an Application to Perceptual Grouping}
\author {
    Guangyao Zhou,
    Wolfgang Lehrach, 
    Antoine Dedieu, 
    Miguel Lázaro-Gredilla,
    Dileep George\\
}
\affiliations {
    Vicarious AI \\
    \{stannis, wolfgang, antoine, miguel, dileep\}@vicarious.com
}

\usepackage{bibentry}

\begin{document}

\maketitle

\begin{abstract}
	Discrete undirected graphical models, also known as Markov Random Fields (MRFs), can flexibly encode probabilistic interactions of multiple variables, and have enjoyed successful applications to a wide range of problems. However, a well-known yet little studied limitation of discrete MRFs is that they cannot capture context-specific independence (CSI). Existing methods require carefully developed theories and purpose-built inference methods, which limit their applications to only small-scale problems. In this paper, we propose the Markov Attention Model (MAM), a family of discrete MRFs that incorporates an \emph{attention} mechanism. The \emph{attention} mechanism allows variables to dynamically \emph{attend to} some other variables while \emph{ignoring} the rest, and enables capturing of CSIs in MRFs. A MAM is formulated as an MRF, allowing it to benefit from the rich set of existing MRF inference methods and scale to large models and datasets. To demonstrate MAM's capabilities to capture CSIs at scale, we apply MAMs to capture an important type of CSI that is present in a symbolic approach to recurrent computations in perceptual grouping. Experiments on two recently proposed synthetic perceptual grouping tasks and on realistic images demonstrate the advantages of MAMs in sample-efficiency, interpretability and generalizability when compared with strong recurrent neural network baselines, and validate MAM's capabilities to efficiently capture CSIs at scale.
\end{abstract}

\section{Introduction}

\textbf{Context-specific independence\quad}Probabilistic Graphical Models (PGMs) compactly encode conditional independence information in multivariate distributions with graph-based representations. However, a well-known limitation of PGMs is that they cannot capture context-specific independence (CSI)~\cite{boutilier1996context, friedman1998learning, fridman2003mixed, nyman2014stratified}, i.e. independence relationships that only hold for certain joint variable states. Many past works~\cite{boutilier1996context, friedman1998learning, poole2003exploiting} focus on encoding CSIs for directed graphical models, also known as Bayesian Networks (BNs). Discrete undirected graphical models, also known as Markov Random Fields (MRFs), have enjoyed successful applications to a wide range of problems~\cite{koller2009probabilistic}. Yet exploiting CSI in discrete MRFs has received relatively little attention. Existing methods typically augment MRFs, e.g.\ by introducing special node-valued random variables~\cite{fridman2003mixed}, or by annotating graph edges~\cite{nyman2014stratified}. However, the augmented models are no longer MRFs, and as a result require carefully developed theories and purpose-built inference methods. The lack of standard tooling has so far limited their applications to only small-scale problems.

In this paper, we propose the Markov Attention Model (MAM), a family of discrete MRFs that incorporates an \emph{attention} mechanism to flexibly capture CSIs in undirected models. The \emph{attention} mechanism allows variables to dynamically (i.e.\ in a context-specific way) \emph{attend to} (i.e.\ conditionally depend on) some other variables while \emph{ignoring} (i.e.\ being conditionally independent of) the rest. We implement the attention mechanism with higher-order factors (HOFs) and discrete attention variables. Each attention variable is connected to two regular variables, and encodes a) information about the joint states of the two connected regular variables with a set of \texttt{ON} states; and b) CSI between the two regular variables with a special \texttt{OFF} state. The HOFs enforce consistency between \texttt{ON} attention variables and their connected regular variables, and capture CSIs by turning certain attention variables \texttt{OFF} for given regular variable states.

Despite the introduction of attention variables, existing MRF inference methods are readily applicable to MAMs due to its formulation as a discrete MRF. In our experiments, we leverage this fact to develop efficient inference methods for MAMs based on max-product belief propagation (MPBP)~\cite{weiss2001optimality}, resulting in orders of magnitude improvements in scalability.  MAM provides a general and efficient approach for capturing CSIs using MRFs, with potential applications to myriad discrete data, e.g.\ gene regulatory networks~\cite{shmulevich2002probabilistic}, word sense disambiguation~\cite{navigli2009word}, and a symbolic approach to recurrent computations for perceptual grouping.

\noindent\textbf{Perceptual grouping and recurrent computations\quad}To demonstrate the capabilities of MAM at scale, we focus on modeling recurrent computations in perceptual grouping using MAM. Perceptual grouping puts multiple parts into a whole, and is a fundamental component of visual intelligence. Inspired by evidence~\cite{roelfsema2006cortical, roelfsema2011incremental, Gilbert2013-co} on the importance of top-down and lateral \emph{recurrent connections}~\cite{Lamme1998-di,Sporns2004-lt} for perceptual grouping, recent computational models~\cite{linsley2018learning, Kim2020Disentangling} incorporate \emph{recurrent computations} using recurrent neural networks (RNNs), and achieve state-of-the-art (SOTA) performance when compared with purely feedforward convolutional neural networks (CNNs). However, as we show in our experiments, they suffer from the now well-known limitations of deep neural networks (DNNs)~\cite{serre2019deep, marcus2018deep}: they require a large amount of training data, are hard to interpret, and do not generalize well to novel setups.

It has long been postulated~\cite{hinton1990preface, bader2005dimensions, garnelo2019reconciling, marcus2020next, garcez2020neurosymbolic} that an effective approach for overcoming such limitations involves high-level, abstract reasoning with symbol-like entities (e.g. abstract \emph{objects} and \emph{object parts}). In this paper, we take such a symbolic approach, and build probabilistic models of abstract objects/object parts and their interactions. We identify an important type of CSI that is present in such models: object parts are conditionally independent when they belong to different objects. We show how we can formulate MAMs to flexibly capture such CSIs and build compact probabilistic models that are sample-efficient, interpretable and generalizable.

\noindent\textbf{Scope\quad}Our MAMs for perceptual grouping operate on the symbolic level, and as a result face the symbol grounding problem~\cite{harnad1990symbol}. Since our goal is to demonstrate MAM's capabilities to capture CSIs at scale, we experiment with two recently proposed synthetic perceptual grouping tasks with binary images, the pathfinder challenge and the cABC challenge~\cite{linsley2018learning, Kim2020Disentangling}, for which we can solve the symbol grounding problem with a \emph{sparsifier}. A sparsifier is a simple binary MRF. It connects the perceptual representations (the binary images) and the symbolic representations (the object parts), and enables learning of MAMs directly from binary images (e.g. \textbf{Fig.~\ref{fig:pathfinder_decoding}}).

Although the two perceptual grouping tasks are synthetic in nature, they are by no means simple. In fact, sophisticated CNN baselines using ResNet~\cite{ronneberger2015u} and U-net~\cite{ronneberger2015u} struggle on these challenges, and novel RNNs are needed to satisfactorily solve the challenges~\cite{Kim2020Disentangling}. In addition, SOTA transformers~\cite{tay2020long, xiong2021nystr, lee2021fnet, moskvichev2021updater} were recently applied to lower-resolution variations of the pathfinder challenge, but perform poorly when compared with MAMs or our RNN baselines~\cite{Kim2020Disentangling}. Nevertheless, as a proof-of-concept, we show that, with some manual design, we can easily apply MAMs trained with rich prior information on singulated objects in the form of object masks to realistic images. The additional experiment hints at MAM's potential in fields like robotics~\cite{sunderhauf2018limits}, where we benefit from structured probabilistic models that can be trained in a sample-efficient way while taking into account available rich prior information (e.g.\ 3D object models). However, MAM is not intended as a replacement for existing methods in situations with much training data but no rich prior information.

\noindent\textbf{Contributions\quad}The main contributions of the paper are:

\begin{enumerate}
	\item We propose MAM, a family of discrete MRFs that flexibly captures CSIs in MRFs. Since a MAM is formulated as a discrete MRF, no additional theory is needed, and existing MRF inference methods are readily applicable. This allows MAMs to scale to large models and datasets.
	\item We identify an important type of CSI in a symbolic approach to recurrent computations for perceptual grouping, and use this as a test case to demonstrate MAM's capabilities to capture CSIs in MRFs at scale.
	\item We apply MAMs to two recently proposed synthetic perceptual grouping tasks. We develop efficient learning of MAMs directly from binary images using large-scale datasets (e.g. with $900$K $300\times 300$ images), and scalable approximate inference procedures based on MPBP to test large MAMs. Experimental results demonstrate MAMs' advantages when compared with SOTA RNN baselines: a) MAMs are more sample-efficient, and order(s) of magnitude more so in challenging situations, b) MAMs give interpretable and semantically meaningful decodings, c) MAMs exhibit high generazibility across setups, while RNNs struggle or completely fail to generalize.
	\item We additionally demonstrate MAMs' power and flexibility by applying MAMs trained with rich prior information on singulated objects to realistic images. 
\end{enumerate}

\noindent\textbf{Presentation style\quad}For readibility, we focus on intuitive descriptions in the main text, and defer the formal, rigorous description of all models and methods to the supplementary.

\section{Background and related works}

\noindent\textbf{Context-specific independence\quad}For BNs, CSI brings more compact encoding and significant inference speedups~\cite{boutilier1996context}, and allows learning of BN models that better emulate the real complexity of the interactions present in the data~\cite{friedman1998learning}. However, exploiting CSI for MRFs has received relatively little attention~\cite{fridman2003mixed, nyman2014stratified}. As a simple example for CSI in MRFs, consider the following gene network from~\cite{fridman2003mixed}: 
\begin{displayquote}
	The protein product pA of gene A induces expression of gene B. Compound C, when present, modifies pA. The modified protein is unable to regulate B directly but can induce expression of gene D. Genes B and D are corepressive (pB inhibits expression of D; pD inhibits expression of B).
\end{displayquote}
Denote by $X_A, X_B, X_D \in \left\{ 0, 1 \right\} $ the expression levels of genes A, B, D, and by $X_C \in \left\{ 0, 1 \right\} $ the presence/absence of compound C. The network contains obvious CSIs: $X_A, X_B$ are only conditionally independent when $X_C = 1$ while $X_A, X_D$ are only conditionally independent when $X_C = 0$. Despite such CSIs, \cite{fridman2003mixed} shows that every pair of variables in the MRF modeling this network are directly dependent, and the MRF is \emph{fully connected} (\textbf{Fig.~\ref{fig:gene_network}[left]}).

In~\cite{fridman2003mixed}, the author captures CSIs by replacing $X_C$ with a node-valued random variable taking values in  $\left\{ X_B, X_D \right\} $ to represent which gene is directly regulated by A, but has to carefully establish the Markov property and develop inference methods based on a customized Gibbs sampler. In~\cite{nyman2014stratified}, the authors annotate graph edges to capture CSIs, but the theory and application are restricted to a simplistic family of \emph{decomposable} models.

\noindent\textbf{Perceptual grouping\quad}Probabilistic methods for problems related to perceptual grouping have a rich history in computer vision~\cite{ren2005cue, ren2006figure, hoiem2007recovering, he2010occlusion, silberman2014contour}. They usually model \emph{concrete entities} (e.g. boundaries and surfaces) using RGB and depth information. In this paper, we explore an alternative, symbolic approach where we build probabilistic models of abstract objects/object parts and their interactions. This symbolic approach reveals an important type of CSI, making it a suitable test case for MAM.

\noindent\textbf{Recurrent computations using RNNs\quad}Motivated by the abundant recurrent connections in the visual cortex, several recent computational models~\cite{Kubilius2019-up, Liang2015-ks, liao2016bridging, spoerer2017recurrent, zamir2017feedback, nayebi2018task, linsley2018learning, Kim2020Disentangling} incorporate recurrent computations using RNNs. They outperform purely feedforward CNNs with more compact models/fewer parameters, yet suffer from well-known limitations of DNNs in sample-efficiency, interpretability and generalizability. There have been significant efforts in the RNN literature on addressing the above limitations. In particular, several recent works~\cite{goyal2020recurrent, mittal2020learning, goyal2021neural} explore ideas similar to our identified CSI for perceptual grouping for better generalizability. In this paper, we demonstrate how MAM can implement such CSI within an MRF, and obtain better sample-efficiency and interpretability, in addition to improved generalizability.

\noindent\textbf{Attention\quad}Attention was first introduced in deep learning~\cite{bahdanau2014neural, mnih2014recurrent}. It has subsequently found great success in both natural language processing~\cite{vaswani2017attention, galassi2020attention} and computer vision~\cite{han2020survey, khan2021transformers}, and plays an important role in many graph neural networks~\cite{velivckovic2017graph, lee2019attention}. In this paper, we reinterpret attention as a mechanism for implementing CSIs in MRFs, and demonstrate its usefulness in symbolic recurrent computations for perceptual grouping.

\noindent\textbf{Max-product belief propagation\quad}MPBP is a message-passing based inference method for approximately computing the maximum-a-posteriori (MAP) state in a PGM. It is highly scalable since it involves only parallelizable local computations. We use MPBP in the log domain with damping~\cite{cohen2017max} for inference in all our models.

\noindent\textbf{Compositional object models\quad}Our symbolic approach to perceptual grouping is similar in spirit to compositional object models, i.e.\ hierarchical probabilistic models of objects/object parts. Some past works~\cite{jin2006composition, zhu2007stochastic, lake2015human} rely on less scalable Markov Chain Monte Carlo or heuristic methods for inference. Others~\cite{george2017generative, Felzenszwalb2020grammar} similarly adopt PGMs with belief propagation for inference. Among these, the recursive cortical network~\cite{george2017generative} is closely related to our object-specific MAM for cABC, but lacks the capacity for object-agnostic modeling (e.g. for pathfinder). Probabilistic scene grammars~\cite{Felzenszwalb2020grammar} model top-down interactions using grammar production rules, but lack proper modeling of lateral interactions.

\section{Methods}

\subsection{Markov Attention Model}
\begin{figure}[t!]
	\centering
	\includegraphics[width=0.47\textwidth]{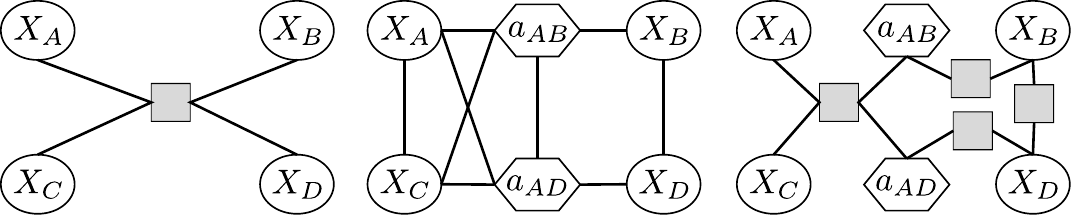}
	\caption{MRFs for the gene network from~\cite{fridman2003mixed}. [Left] Factor graph for the MRF modeling the network. The shaded square represents the single HOF. [Center] Graph of conditional independence relationships for the MAM modeling the network. [Right] Factor graph for the MAM modeling the network. Shaded squares represent factors.}
	\label{fig:gene_network}
\end{figure}

\begin{figure*}[t!]
	\centering
	\includegraphics[width=\textwidth]{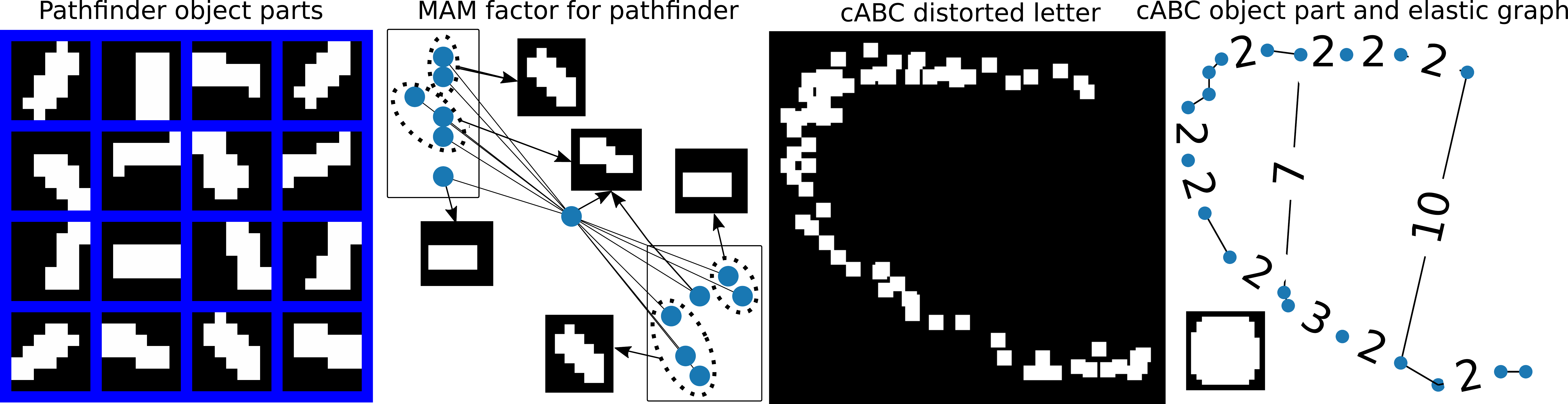}
	\caption{[Left] $16$ learned object parts for the pathfinder challenge. [Center left] Top $12$ (out of more than 800) possible contour continuations for a given object part. Blue dots represent object parts. Each rectangle represents contour continuations on one side. Dotted circles represent the same type of object parts at different locations. Arrows point to visualizations of the corresponding object parts. [Center right] A noisy, distorted letter from the cABC challenge. [Right] A single learned object part for the cABC challenge, and the extracted elastic graph for the center right letter. Visualized numbers are the perturb radiuses associated with the edges (elastic constraints). Edges with no associated numbers have perturb radius $1$.}
	\label{fig:model_illustration}
\end{figure*}

\noindent\textbf{Revisiting the gene network example\quad}We start by showing how we can formulate a MAM to encode the obvious CSIs in the gene network example from~\cite{fridman2003mixed}. We adopt the standard factor graph (FG) representation~\cite{kschischang2001factor} for discrete MRFs. An FG is a bipartite graph connecting variables to factors. 

To capture CSIs, we introduce two binary attention variables $a_{AB}, a_{AD}\in \left\{ \text{\texttt{ON}, \texttt{OFF}} \right\} $, one for each pair of variables with CSI. Intuitively, the attention variables indicate whether A directly regulates B, D. The MAM modeling the gene network is a discrete MRF on 4 regular variables $X_A, X_B, X_C, X_D$ and the two introduced attention variables. We visualize its graph in \textbf{Fig.~\ref{fig:gene_network}[center]} and its FG in \textbf{Fig.~\ref{fig:gene_network}[right]}. The MAM is specified by 4 factors:
\begin{enumerate}
	\item A HOF involving $X_A, X_C, a_{AB}, a_{AD}$. It models A, C interactions and captures the CSIs with 4 joint states:
		\begin{enumerate}
			\item $X_A=0, X_C=0, a_{AB}\text{ is \texttt{OFF}}, a_{AD}\text{ is \texttt{OFF}}$.
			\item $X_A=0, X_C=1, a_{AB}\text{ is \texttt{OFF}}, a_{AD}\text{ is \texttt{OFF}}$.
			\item $X_A=1, X_C=0, a_{AB}\text{ is \texttt{ON}}, a_{AD}\text{ is \texttt{OFF}}$.
			\item $X_A=1, X_C=1, a_{AB}\text{ is \texttt{OFF}}, a_{AD}\text{ is \texttt{ON}}$.
		\end{enumerate}
		Joint states (a), (b) encode A, C interactions, while joint states (c), (d) capture the CSIs.
	\item A factor involving $a_{AB}, X_B$. It promotes $X_B=1$ when A directly regulates B ($a_{AB}$ is \texttt{ON}), and gives 0 potential (in log domain) when A does not ($a_{AB}$ is \texttt{OFF}).
	\item A factor involving $a_{AD}, X_D$, similar to the above.
	\item A factor involving $X_B, X_D$, modeling corepressiveness.
\end{enumerate}

The \texttt{ON} state of an attention variable encodes information about joint states of its connected regular variables (e.g. $a_{AB}$ is \texttt{ON} means $X_A=1$ and $X_B$ is likely $1$). Factors involving regular and attention variables enforce consistency between regular variables and \texttt{ON} attention variables (e.g. factor 1 enforces $X_A=1$ when any connected attention variable is \texttt{ON}, and factor 2 enforces $X_B$ is likely  $1$ when  $a_{AB}$ is \texttt{ON}), and models CSIs by turning certain attention variables \texttt{OFF} (e.g. joint states (c), (d) in factor 1). In what follows, we present the general formulation of MAM, and demonstrate in later sections how we can formulate, learn and apply MAMs to capture CSIs, with perceptual grouping as an example.

\noindent\textbf{General formulation\quad} Given a FG, we say two variables are $\psi$-connected if there exists a factor that connects to both variables. A MAM is defined as a discrete MRF with two types of variables and additional connectivity constraints:
\begin{enumerate}
	\item A variable is either a regular or an attention variable.
	\item An attention variable is $\psi$-connected to exactly two regular variables. We  denote the attention variable that is $\psi$-connected to the regular variables $x_u, x_v$ by $a_{\left\{ x_u, x_v \right\} }$.
	\item A factor that is connected to an attention variable is also connected to exactly one regular variable that is $\psi$-connected to the attention variable.
\end{enumerate}
For the attention variables, we additionally designate \texttt{ON} states encoding information about joint states of the two $\psi$-connected regular variables, and special \texttt{OFF} states encoding CSIs. The factors model regular variable interactions, enforce consistency between regular and attention variables, and capture CSIs with \texttt{OFF} attention variables.


\begin{figure*}[t!]
	\centering
	\includegraphics[width=\textwidth]{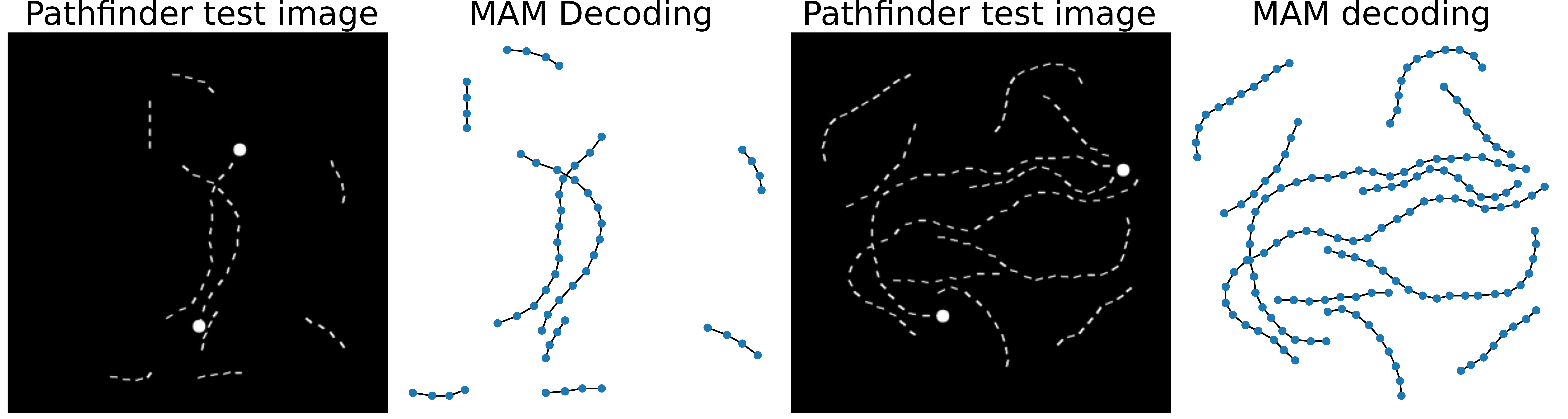}
	\caption{[Left] A \emph{hard} pathfinder test image. [Center left] MAM decoding for the left image. [Center right] An \emph{ultimate} pathfinder test image. [Right] MAM decoding for the center right image.}
	\label{fig:pathfinder_decoding}
\end{figure*}

\begin{table*}[h!]
    \centering
    \caption{\small{Generalization across difficulties on pathfinder for H-CNN (trained on $900$K images) and MAM (trained on $5$K images). We follow~\cite{Kim2020Disentangling} for training H-CNN on \emph{easy, medium} and \emph{hard}, but train for $4\times$ longer on \emph{ultimate}. MAM can easily generalize across difficulty levels, while H-CNN completely fails to generalize, despite being trained for longer and on $180\times$ more images. We expect similarly good performances from MAMs trained on \emph{ultimate}, but did not retrain since we only introduce \emph{ultimate} as an additional generalization test.}}
    \resizebox{.85\textwidth}{!}{
   \begin{tabular}{ccccccccc}
    \toprule
    \multirow{2}{*}{\diagbox{Train}{Test}}& \multicolumn{2}{c}{Easy} & \multicolumn{2}{c}{Medium} & \multicolumn{2}{c}{Hard} & \multicolumn{2}{c}{Ultimate} \\
    & H-CNN & MAM & H-CNN & MAM & H-CNN & MAM  & H-CNN & MAM\\
    \midrule
    Easy   &  \textbf{99.47\%} & 98.74\% & 49.01\% & \textbf{98.65\%} & 49.48\% & \textbf{98.21\%} & 49.88 \% & \textbf{97.21\%}\\
    Medium &  45.71\% & \textbf{98.92\%} & 98.02\% & \textbf{99.00\%} & 49.55\% & \textbf{98.77\%}  & 49.90 \% & \textbf{98.33\%}\\
    Hard  &  46.95\% & \textbf{98.92\%} & 48.52\% & \textbf{99.07\%} & 97.60\% & \textbf{98.80\%}  & 49.94 \% & \textbf{98.42\%} \\
    Ultimate & 29.52\% & -- & 39.67\% & -- & 56.94\% & -- & 79.68\% & -- \\
    \bottomrule
    \end{tabular}
    \label{tab:pathfinder}
    }
\end{table*}

\subsection{Applying MAMs to perceptual grouping} 
We take a symbolic approach to perceptual grouping, and build probabilistic models of abstract objects/object parts and their interactions. Inspired by the importance of top-down and lateral recurrent connections, we focus on implementing two types of \emph{symbolic recurrent computations}: a) modeling part-whole relationships with top-down parent/children interactions, and b) modeling object parts co-occurrences with lateral contextual interactions. We introduce binary \emph{part variables} as the regular variables in our MRFs to model the presence/absence of objects/object parts.

We consider generic and reusable object parts, i.e.\ object parts that can belong to many different objects, and interact with many object parts. Modeling such interactions with an MRF on part variables quickly becomes infeasible due to large HOFs with exponential number of states. However, we observe an important type of CSI: object parts are conditionally independent when they belong to different objects. We can exploit such CSI to build more compact models. For example, if we know an object part interacts with two other object parts (e.g.\ as part of a closed contour), after identifying which two object parts it interacts with, it becomes independent of any other object parts.

We can easily formulate MAMs to capture such CSI. We introduce binary attention variables, one for each pair of interacting part variables. An \texttt{ON} attention variable implies both connected part variables are \texttt{ON}, while an \texttt{OFF} attention variable encodes CSI between the two connected part variables. We introduce HOFs, one for each part variable. A HOF involves its associated part variable and all its connected attention variables. It enforces consistency between the part and attention variables: the part variable should be \texttt{ON} when at least one of its connected attention variable is \texttt{ON}. It encodes CSI by turning \texttt{ON} a sparse (but non-empty) set of attention variables when the part variable is \texttt{ON}. The sparse set of \texttt{ON} attention variables models both part-whole relationships and object parts co-occurences, while also serves as the context for CSI with other part variables.

We additionally incorporate invariance by partitioning the set of attention variables connected to a part variable. Each subset models variations in location of an interacting object part. A HOF identifies the sparse set of interactions by picking first a few subsets (a \emph{prototypical interaction pattern}) followed by exactly one attention variable within each subset. See the supplementary for a formal description. 

By encoding CSIs, MAMs can focus on modeling essential interactions within single objects, without worrying about multi-object interactions. As we see in our experiments, this translates into compact probabilistic models with superior sample-efficiency and generalizability. MAMs are additionally highly interpretable because of intuitive formulations: from the result of MAP inference, we can read off which objects/object parts are present from \texttt{ON} part variables, and identify which object parts belong to the same object from connectivity patterns with \texttt{ON} attention variables.

\subsection{Symbol grounding with the sparsifier}\label{sec:sparsify}

A sparsifier is a binary MRF containing two different types of variables: a) binary part variables (same as in MAMs) for the presence/absence of object parts, and b) binary pixel variables for pixels in a binary image. We describe an object part by a set of pixels on the image, and consider the object part as a \emph{parent} of its covered pixels. There is one \texttt{OR} factor per pixel variable, which is a HOF that connects the pixel variable and all its parent part variables. The \texttt{OR} factor implements \emph{explaining away}~\cite{wellman1993explaining}: the pixel is most likely \texttt{ON} when at least one parent is \texttt{ON}, and most likely \texttt{OFF} when all parents are \texttt{OFF}. See the supplementary for a formal description of the model.

\noindent\textbf{Connecting MAMs with binary images\quad}In our experiments, we combine a sparsifier with a MAM to form a probabilistic model for binary images: given a MAM and a binary image, we associate with each part variable in the MAM a set of pixels in the binary image, and combine the HOFs in the MAM with the \texttt{OR} factors in the sparsifier.

\noindent\textbf{Sparsification\quad}If we bias each part variable towards \texttt{OFF}, because of \emph{explaining away}, we can use inference with the sparsifier to \emph{sparsify} a binary image, and extract symbolic representations (a sparse set of object parts activations) from perceptual representations (dense binary images). See the supplementary for an illustrative example.

\noindent\textbf{Learning MAMs directly from images\quad}As a special case of~\cite{lazarogredilla2017hierarchical}, we can augment a sparsifier to represent the object parts as a set of binary weight variables. An \texttt{ON} weight variable means the object part covers the corresponding pixel. We can then use joint inference on the part, pixel and weight variables over multiple images, with the weight variables shared convolutionally and across images, to learn object parts directly from images without supervision. In our experiments, we learn MAMs by first learning the object parts on a few training images using the above procedure, before sparsifying training images with the learned object parts to learn the relevant interactions.

\section{Experiments}

We develop efficient MPBP updates for the HOFs in MAMs and the sparsifier by exploting their simple structures, and implement MPBP using \emph{JAX}~\cite{jax2018github} with parallel updates of all messages at each iteration to fully leverage the power of modern accelerators like GPUs. See the supplementary for detailed message updating equations. Code is available at \url{https://github.com/vicariousinc/mam/}.

\begin{figure*}[t!]
	\centering
	\includegraphics[width=\textwidth]{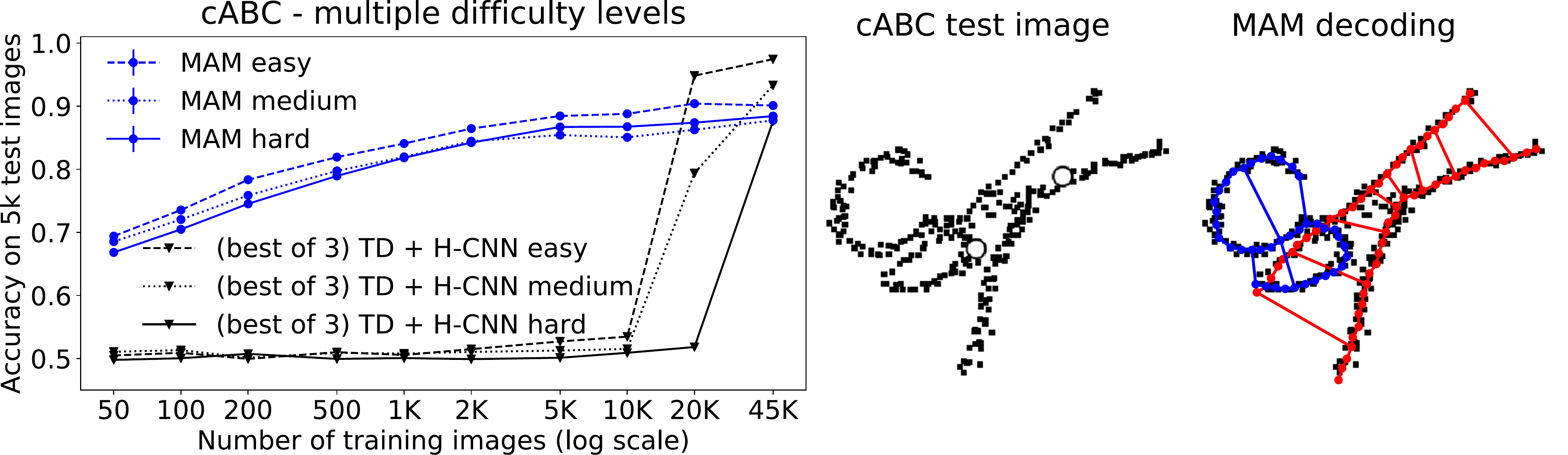}
	\caption{\small{[Left] Evolution of classification accuracy on 5K test images for MAM and TD+H-CNN as the number of training images increase. We average results over three runs for MAMs, but follow~\cite{Kim2020Disentangling} and report the best result over three runs for TD+H-CNN. Error bars are included for MAMs, but are barely visible due to small accuracy variations across different runs. TD+H-CNN completely fails with less than $20$K training images, while MAM performs well above chance with only $50$ training images. [Center]  cABC test image. [Right] MAM decoding of the center image, showing two elastic graphs matched to the letters.}}
	\label{fig:cabc}
\end{figure*}

\begin{table*}[h!]
    \centering
    \caption{\small{Generalization across difficulties on cABC for TD+H-CNN and MAM (both trained on $45$K images).}} 
    \resizebox{.85\textwidth}{!}{
   \begin{tabular}{ccccccc}
    \toprule
    \multirow{2}{*}{\diagbox{Train}{Test}}& \multicolumn{2}{c}{Easy} & \multicolumn{2}{c}{Medium} & \multicolumn{2}{c}{Hard} \\
    & TD+H-CNN & MAM & TD+H-CNN & MAM & TD+H-CNN & MAM\\
    \midrule
    Easy   & \textbf{97.50\%} &  90.10\% & \textbf{89.60\%} & 87.40\% & 57.90\% & \textbf{86.36\%} \\
    Medium &  \textbf{95.00\%} & 89.22\% & \textbf{93.00\%} & 87.74\% & 61.60\% & \textbf{86.82\%}\\
    Hard  &  82.40\% & \textbf{89.00\%} & 82.20\% & \textbf{87.20\%} & 85.60\% & \textbf{88.42\%} \\
    \bottomrule
    \end{tabular}
    \label{tab:cabc}
    }
\end{table*}

\begin{figure}[t!]
    \centering
    \includegraphics[width=.47\textwidth]{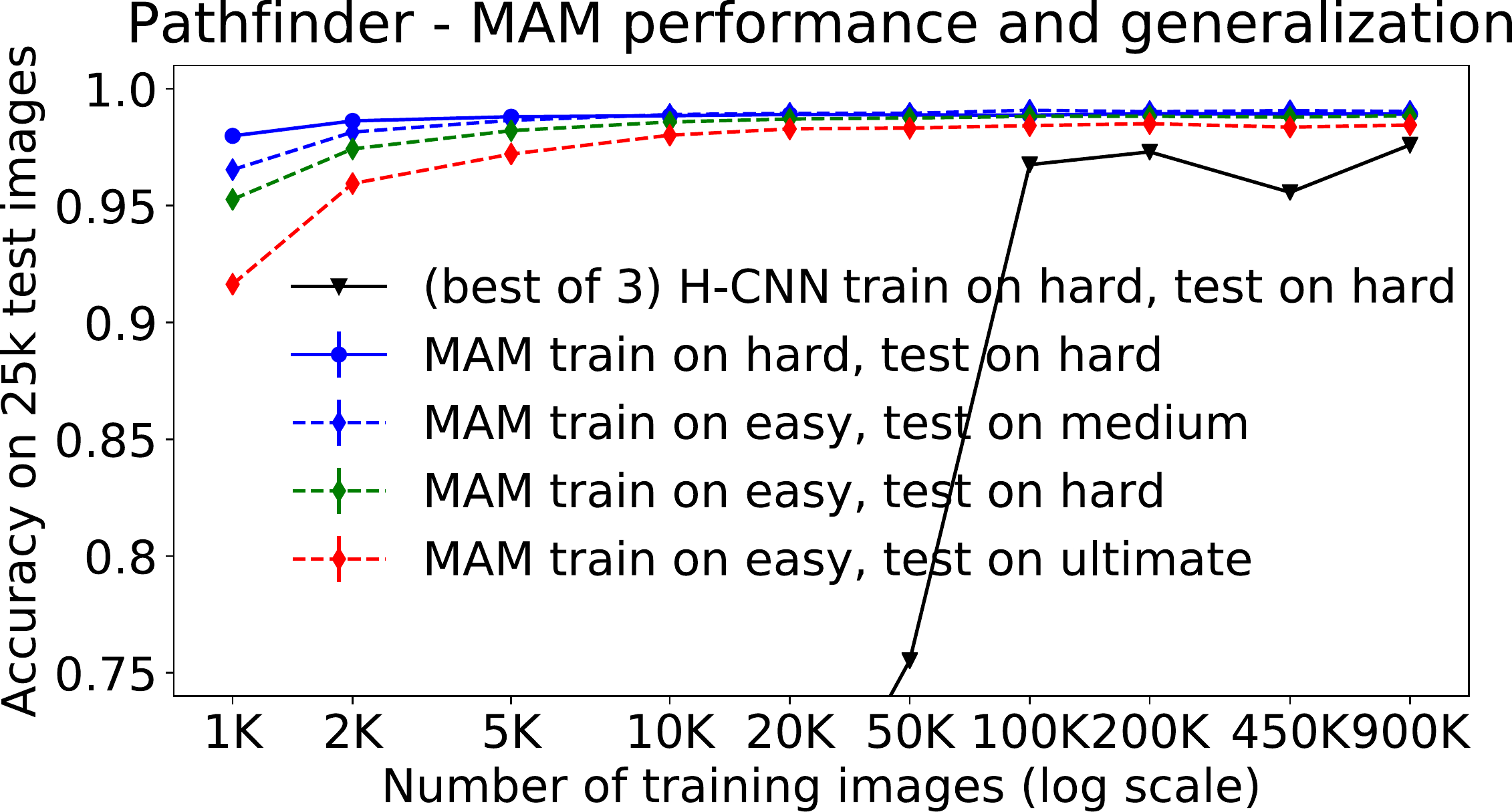}
    \caption{\small{MAM is more sample-efficient than H-CNN on \emph{hard}. MAMs generalize well across difficulties, even with a small number of training images. Notably, an MAM trained on 1K \emph{easy} images gets $91\%$ accuracy on \emph{ultimate}, while an H-CNN trained for $4\times$ longer than \cite{Kim2020Disentangling} on $900$K \emph{ultimate} images achieves less than $80\%$ accuracy on \emph{ultimate} (Tab.~\ref{tab:pathfinder}).}}
    \label{fig:pathfinder_accuracy}
\end{figure}

\subsection{Pathfinder challenge}\label{sec:pathfinder}

\noindent\textbf{Task\quad}The pathfinder challenge~\cite{linsley2018learning, Kim2020Disentangling} uses multiple broken contours as objects, and involves identifying whether two markers are on the same object (\textbf{Fig.~\ref{fig:pathfinder_decoding}}). It is expected to be solved by an object-agnostic recurrent routine that groups neighboring low-level visual features according to Gestalt laws. We follow the setups in~\cite{Kim2020Disentangling}, and use the proposed H-CNN, a SOTA RNN that models lateral interactions, as our baseline. 

\noindent\textbf{Embedding Gestalt laws in an object-agnostic MAM\quad}We specify a MAM with $16$ types of object parts (\textbf{Fig.~\ref{fig:model_illustration}[left]}), each with possible presence in an $M\times N$ grid of locations. We extract co-occurences of object parts from data (see \textbf{Learning}). Each co-occurence is applied convolutionally to define a set of attention variables. For a part variable, we partition its associated attention variables into $2$ sets to capture possible contour continuations on both sides (\textbf{Fig.~\ref{fig:model_illustration}[center left]}). A HOF is specified by a) a set of prototypical interaction patterns enforcing that a contour should continue on one or both sides, and b) the potential in log domain, which penalizes contour termination and encourages the contour to be \texttt{OFF} or continue on both sides. We connect the MAM to the image with a sparsifier to model the presence of a variable number of broken contours.

\noindent\textbf{Learning\quad}We can fully specify the object-agnostic MAM by extracting a set of co-occurences of object parts from data. We achieve this by sparsifying individual broken contours (available as part of the data generation process) with the sparsifier, using $16$ object parts learned from 10 images(\textbf{Fig.~\ref{fig:model_illustration}[left]}). For each individual broken contour, we obtain sparse object parts activations from MPBP decoding, and heuristically derive the co-occurences of object parts. 

\noindent\textbf{Inference\quad}Due to the clean data in the pathfinder challenge, we aggressively prune part and attention variables, by sparsifying a given test image and keeping only the activated part variables and the associated attention variables. We apply MPBP to the pruned MAM for decoding and classification. For images with well-separated contours, this pruning alone can already interpret and classify the images well. However, subsequent inference with MAM helps develop valid interpretations in challenging situations with close or crossing contours (e.g. \textbf{Fig.~\ref{fig:pathfinder_decoding}}), and improve classification.

\noindent\textbf{Results\quad}We follow~\cite{Kim2020Disentangling} and experiment with $3$ difficulty levels, \emph{easy, medium, hard}. Although~\cite{Kim2020Disentangling} used 900K images for all levels, empirically we find that H-CNN can achieve good performance with 1K images for \emph{easy}, and 2K images for \emph{medium}. On \emph{hard}, the most interesting level, \textbf{Fig.~\ref{fig:pathfinder_accuracy}} shows the additional sample-efficiency of MAM compared to H-CNN: H-CNN struggles even with 50K training images, while MAM achieves close to perfect accuracy with 1K training images. MAM additionally provides semantically meaningful interpretations of the image as multiple contours (e.g. \textbf{Fig.~\ref{fig:pathfinder_decoding}}).
\textbf{Fig.~\ref{fig:pathfinder_accuracy}} and \textbf{Tab.~\ref{tab:pathfinder}} show MAM's superior generalizability. MAMs trained with only $1$K images can easily generalize, while H-CNNs trained on $900\times$ more images still completely fail. As an additional and more interesting generalization test, we introduce a new difficulty level called \emph{ultimate}, with more, longer contours (see \textbf{Fig.~\ref{fig:pathfinder_decoding}[center right]} for an example, and the supplementary for more details). MAM again exhibits strong generalization capacity: a MAM trained on 1K \emph{easy} images already achieves over $91\%$ accuracy on \emph{ultimate}, and MAMs trained with 5K images on any level achieve over $97\%$ accuracy on \emph{ultimate}. In contrast, an H-CNN trained on 900K \emph{ultimate} images for $4\times $ longer than~\cite{Kim2020Disentangling} only achieves less than $80\%$ accuracy on \emph{ultimate}.

\subsection{cABC challenge}\label{sec:cabc}
\noindent\textbf{Task\quad}The cABC challenge~\cite{Kim2020Disentangling} uses a pair of noisy, distorted letters as objects (\textbf{Fig~\ref{fig:cabc}[center]}), and seeks to identify whether two markers are on the same object. It complements the pathfinder challenge in that Gestalt strategies are expected to be ineffective, and it instead needs an object-specific recurrent routine that imposes high-level expectations about the shapes and structures of the letters. We follow the setups in~\cite{Kim2020Disentangling}, and use the proposed TD-H-CNN, a SOTA RNN that models both top-down and lateral interactions, as our baseline.

\noindent\textbf{Elastic graphs and object-specific MAMs\quad}We model \emph{elastic graphs} with object-specific MAMs, and match elastic graphs to noisy, distorted letters to solve the cABC challenge. An elastic graph is an undirected graph that models the letter as a composition of all the object parts represented by its vertices. Each vertex is associated with a reference location, and the object part it represents is allowed move around the reference location. An edge imposes an \emph{elastic constraint} on the two connected object parts, by constraining their distance to be within a \emph{perturb radius} of their reference distance. Elastic graphs use constrained elasticities to model deformations of letters without losing their shapes and structures. Several past works~\cite{lades1993distortion, wiskott1997face, george2017generative} have explored similar ideas. Our formulation and some efficient inference procedures are partly inspired by~\cite{george2017generative}. We extract a set of elastic graphs from data (see \textbf{Learning}), and specify an object-specific MAM for each elastic graph. We merge all individual MAMs into a MAM $\mathcal{M}$, to model the presence of one training letter variation in one location on the image. We connect $2$ copies $\mathcal{M}^{(1)}, \mathcal{M}^{(2)}$ of $\mathcal{M}$ to the image with a sparsifier to model the presence of $2$ noisy, distorted letters in each image. 

\noindent\textbf{Learning\quad}We can fully specify the object-specific MAM by extracting a set of elastic graphs from data. We achieve this by first sparsifying individual letters (available through ground truth segmentations) with the sparsifier, using a single object part learned from 10 images (\textbf{Fig.~\ref{fig:model_illustration}[right]}). We define the vertices in the elastic graph using the sparse object parts activations from MPBP decoding, and heuristically identify a small number of elastic constraints to get the elastic graph (\textbf{Fig.~\ref{fig:model_illustration}[right]}).

\noindent\textbf{Inference\quad}To make inference efficient, we again prune unlikely part variables and unnecessary attention variables by identifying rough locations of a set of best matching elastic graphs, and keep only their associated part and attention variables. We apply heuristic pairwise reasoning to approximate MPBP inference with the sparsifier. These procedures allow us to easily scale to models with $90$K elastic graphs.

\noindent\textbf{Results\quad}We follow~\cite{Kim2020Disentangling} and experiment with $3$ difficulty levels, \emph{easy, medium, hard}. \textbf{Fig.~\ref{fig:cabc}[left]} shows the additional sample-efficiency of MAM compared to TD+H-CNN. TD+H-CNN completely fails with less than $20$K training images, while MAM performs well above chance with only $50$ training images. With $45$K images, MAM performs worse than TD+H-CNN on \emph{easy} and \emph{medium}, but better on the more challenging and interesting \emph{hard} level (see \textbf{Fig.~\ref{fig:cabc}[center]} for an example), and additionally gives interpretable and semantically meaningful elastic graphs matchings (\textbf{Fig.~\ref{fig:cabc}[right]}).
\textbf{Tab.~\ref{tab:cabc}} shows MAM's superior generalizability. TD+H-CNN trained on easier levels struggle to generalize to harder levels. TD+H-CNN trained on \emph{hard} gives decent but lower performance on \emph{easy} and \emph{medium}. In contrast, MAMs trained on any level readily generalize to any other level, and generally exhibit consistent and sensible behavior of decreasing performance as difficulty increases.

\subsection{Proof-of-concept using realistic images}

\noindent\textbf{Task\quad}We use realistic RGB-D images containing $39$ different single objects in totes from the Object Segmentation Training Dataset~\cite{zeng2016multi}\footnote{Available at \url{https://vision.princeton.edu/projects/2016/apc/}}. We randomly sample $70\%$ of the data for training, and use the remaining $30\%$ for testing. We use object masks to filter out images in which the objects are only partially visible. This results in $7797$ training images and $5870$ test images across all objects. We train object-specific MAMs to detect objects in test images.


\noindent\textbf{Learning\quad}We obtain a dense set of edge detections for edges of 16 different orientations, and heuristically sparsify by picking detected edges within the object masks at roughly equal distances to construct elastic graphs.

\noindent\textbf{Results\quad}We predict the object mask as the convex hull of object parts activations from MAM inference, and use the intersection over unions (IOUs) of the predicted object masks and the provided object masks as our performance measure. Note that the number of training images is just $1.3\times$ the number of test images, and objects in test images are in poses unseen in the training images. Despite challenges like deformable objects and adversarial surface properties, we are able to obtain a median IOU of $0.71$ across all objects, demonstrating the sample-efficiency (small number of training images) and generalization capacities (generalizing to unseen novel poses) of the object-specific MAMs trained with rich prior information when applied to realistic images. See the supplementary for some visualizations.



\bibliography{refs}

\end{document}


\maketitle

\tableofcontents

\section{Applying MAMs to perceptual grouping}
\subsection{Formal description}

Our MAMs for perceptual grouping are binary higher-order MRFs containing two different types of variables: (1) binary part variables representing the presence/absence of object parts, and (2) binary attention variables representing the presence/absence of object parts interactions. Each part variable is associated with one HOF, connecting the part variable and its associated attention variables. 

Use $\left| \cdot  \right|$ to denote the cardinality of a set, $\mathcal{P}(\cdot )$ to denote all subsets of a set, and $\mathcal{P}_2(\cdot )$ to denote all cardinality 2 subsets of a set. We can specify our MAM's variables via the interaction graph, which is an undirected graph $\mathcal{G}_I=\left( \mathcal{V}_I, \mathcal{E}_I \right)$ where $\mathcal{V}_I$ represents the set of vertices, and $\mathcal{E}_I = \left\{ \left\{ u, v \right\}: ~ u, v\in \mathcal{V}_I, ~ \exists \text{ an edge in }\mathcal{G}_I \text{ connecting }u, v  \right\} $ represents the set of edges. To  build a MAM from $\mathcal{G}_I$, we associate a binary part variable $x_v$ with each vertex $v \in \mathcal{V}_I$, and a binary attention variable with each edge in $\mathcal{E}_I$. For ease of presentation, in what follows we represent the attention variable associated with $\left\{ u, v \right\} \in \mathcal{E}_I$ interchangeably as either $a_{\left\{ u, v \right\} }$ or $a_{\left\{ x_{u}, x_{v} \right\}}$.

Let $a_{\sim x_v} = \left\{ a_{\left\{ u, v \right\} }: \left\{ u, v \right\} \in \mathcal{E}_I \right\} $ be the set of attention variables associated with the part variable $x_v, v \in \mathcal{V}_I$. The corresponding MAM is specified by $\left| \mathcal{V}_I \right| $ factors $f_{x_v}$, one for each part variable $x_v$ operating on $|a_{\sim x_v}| + 1$ variables (namely $\left\{ x_v \right\} \cup a_{\sim x_v}$). Since all variables are binary, we represent a configuration for $f_{x_v}$ by its entries that are \texttt{ON}, which is an element of $\mathcal{P}\left( \left\{ x_v \right\} \cup a_{\sim x_v} \right)$.

\begin{figure}[t!]
    \centering
    \includegraphics[width=\textwidth]{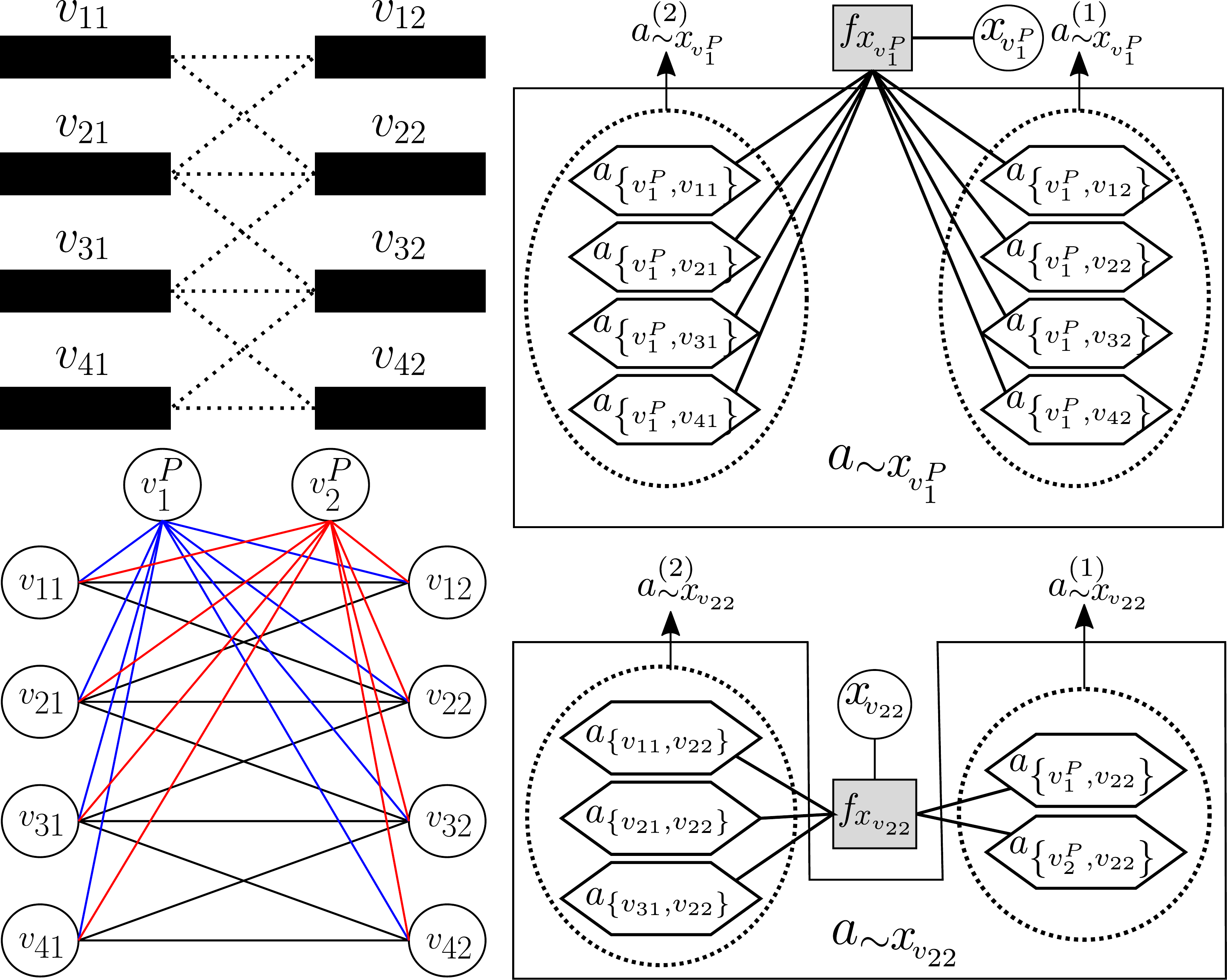}
    \caption{[Upper left] Toy example setup: line segments (solid black rectangles) and their potential continuations (dotted lines). [Lower left] The interaction graph for the corresponding MAM. [Upper right] The HOF $f_{x_{v_1^{P}}}$ modeling top-down interactions. [Lower right] The HOF $f_{x_{v_{2 2}}}$ coordinating top-down and lateral interactions.}
    \label{fig:mam}
\end{figure}

The HOF $f_{x_v}$ is specified by a partition of $a_{\sim x_v}$ into $M_{x_v}$ disjoint subsets of attention variables $a^{(k)}_{\sim x_v}, ~ k=1, \ldots, M_{x_v}$. Intuitively, we can think of each $a_{\sim x_v}^{(k)}$ as an \textit{interaction group}, consisting of a set of interchangeable interactions. The HOF $f_{x_v}$ uses a set of prototypical interaction patterns to control $x_v$'s interaction with other part variables. Each prototypical interaction pattern turns on a set of interaction groups, and the HOF $f_{x_v}$ makes context-specific decisions on which concrete interaction to use within each interaction group.  In particular, this implies that for any valid configuration of the HOF $f_{x_v}$, at most one of the binary variables in each subset $a^{(k)}_{\sim x_v}$ can be \texttt{ON}. See Sec.~\ref{sec:toy} and \textbf{Fig.~\ref{fig:mam}} for a toy example and its HOFs to illustrate the formal descriptions here.

Formally, we specify the HOF $f_{x_v}$ with a set of $M_{x_v}$-dimensional binary vectors $\mathcal{B}_{x_v} \subset \left\{ 0, 1 \right\}^{M_{x_v}} $ where $\B{0} = \begin{pmatrix} 0, \ldots, 0 \end{pmatrix}  \in \mathcal{B}_{x_v} $, and a potential in log domain $\mathcal{U}_{x_v}: \mathcal{B}_{x_v} \mapsto  \mathbb{R} $. Here $\mathcal{B}_{x_v}$ encodes the set of prototypical interaction patterns for $f_{x_v}$. Using $\mathbbm{1}$ to denote the indicator function, each $\B{b} = \left( b_1, \ldots, b_{M_{x_v}} \right)\in \mathcal{B}_{x_v}$ induces a subset of valid configurations $\mathcal{C}_{x_v}(\B{b})$ for the HOF $f_{x_v}$ as
\[
	\mathcal{C}_{x_v}(\B{b}) = \left\{ \B{c} \in \mathcal{P}\left(\left\{ x_v \right\} \cup a_{\sim x_v}\right): \left| \B{c} \cap a_{\sim x_v}^{(k)} \right| = b_k, ~~ k=1, \ldots, M_{x_v} ~~ \text{and} ~~ \mathbbm{1}_{x_v \in \B{c}} = \mathbbm{1}_{\| \B{b} \|_1 > 0} \right\},
\] 
Each $\B{c} \in \mathcal{C}_{x_v}(\B{b})$ fully specifies a joint configuration of the part variable $x_v$ and of all the attention variables $a_{\sim x_v}$. In particular, note that if the part variable $x_v$ is \texttt{ON}, then at least one attention variable in $a_{\sim x_v}$ is \texttt{ON}. Reciprocally, if all the attention variables in $a_{\sim x_v}$ are \texttt{OFF}, then $\B{b}=\B{0}$ and the part variable $x_v$ is \texttt{OFF}, meaning $x_v$ does not interact with and has no effects on other part variables.

With a slight abuse of notation, for a configuration $\B{c} \in \mathcal{P}\left(\left\{ x_v \right\} \cup a_{\sim x_v}\right) $, we use $f_{x_v}\left( \B{c} \right) $ to denote the potential in log domain of $\B{c}$ for $f_{x_v}$. The HOF $f_{x_v}$ adopts a simple structure, and is defined as
\[
	f_{x_v}\left( \B{c} \right) = \begin{cases}
		\mathcal{U}_{x_v} \left( \begin{pmatrix} \left| \B{c} \cap a_{\sim x_v}^{(1)} \right|, \ldots,  \left| \B{c} \cap a_{\sim x_v}^{(M_{x_v})} \right| \end{pmatrix}  \right) , & \forall \B{c} \in \bigcup_{\B{b}  \in \mathcal{B}_{x_v}} \mathcal{C}_{x_v} \left( \B{b}  \right) \\
		-\infty, & \forall \B{c} \not\in \bigcup_{\B{b}  \in \mathcal{B}_{x_v}} \mathcal{C}_{x_v} \left( \B{b}  \right) 
	\end{cases}
\] 

\begin{figure}[h!]
    \centering
    \includegraphics[width=\textwidth]{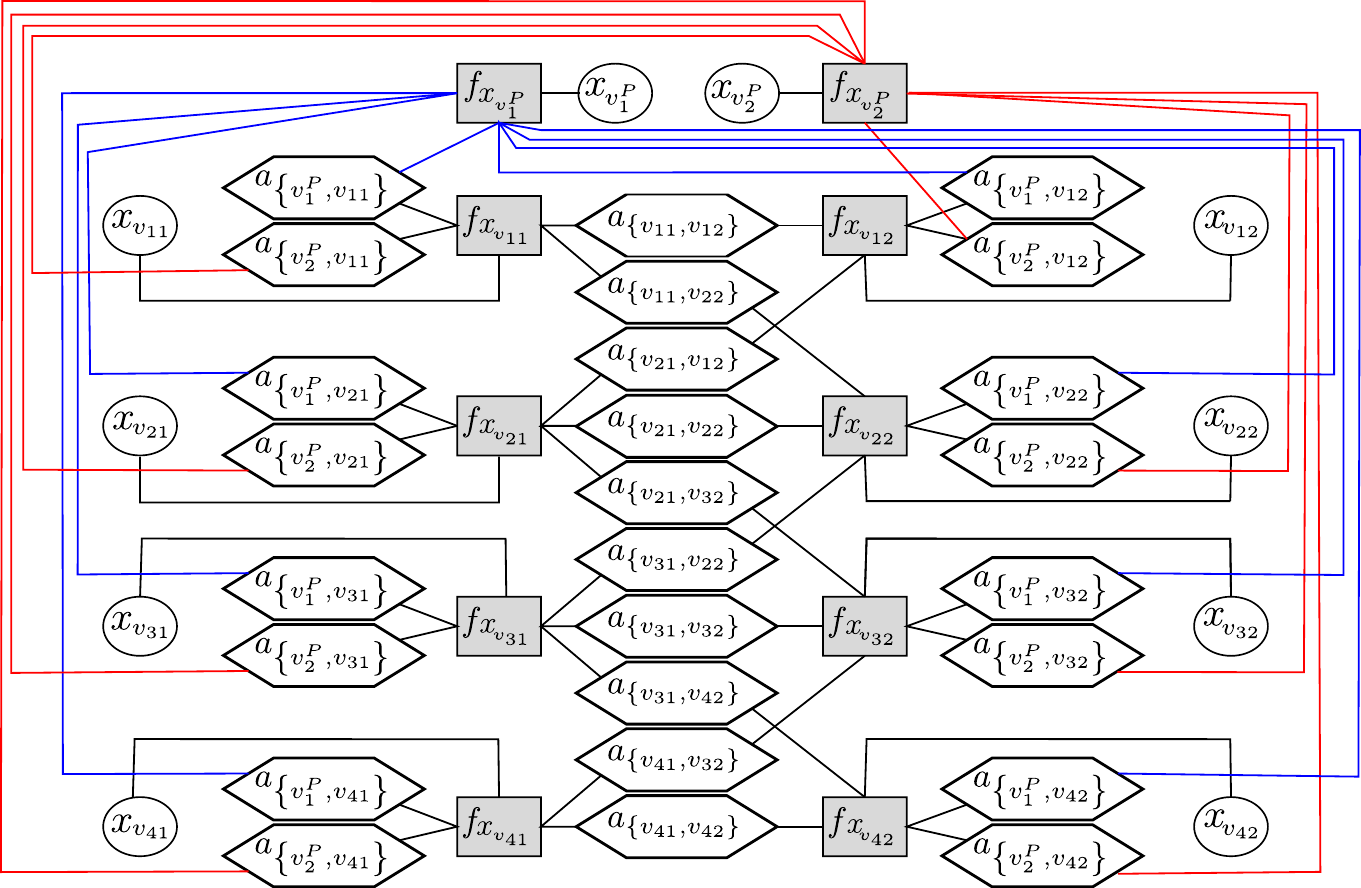}
    \caption{Factor graph for the MAM used in the toy example in Sec. 3.1 in the main text. Circles represent part variables. Hexagons represent attention variables. Shaded squares represent HOFs.}
    \label{fig:factor_graph}
\end{figure}

\subsection{An illustrative toy example}\label{sec:toy}

Our MAMs for perceptual grouping provide a unified framework for symbolic recurrent computations. They coordinately model top-down parent/children interactions and lateral contextual object parts interactions. They are formulated as binary higher-order MRFs, but differ from typical MRFs: MRFs commonly model graph vertices as variables and use graph edges to encode conditional independence, while MAMs model both graph vertices and edges as variables. The attention variables provide a convenient attention mechanism. They allow part variables to selectively pay attention to a small subset of other part variables while ignoring the rest, and enable efficient recurrent computations. In what follows, we work through a toy example to illustrate MAMs' coordinated modeling of both top-down and lateral interactions for perceptual grouping, as well as the efficiency gains from the attention mechanism through attention variables when compared with typical MRFs.

Consider a $4\times 2$ grid of variables $v_{i j}, ~ i=1,\ldots, 4, ~ j=1, 2$. Each variable represents a horizontal line segment that can continue, either in the horizontal direction or right above/below the horizontal direction, to form a line consisting of two segments. See \textbf{Fig.~\ref{fig:mam}[upper left]} for an illustration. We further impose the global constraint that there can be at most two such lines. 

We can easily specify a MAM to model the above. In addition to the $8$ binary part variables $x_{v_{i j}},i=1,\ldots, 4, j=1,2~$, we introduce $2$ additional variables $v^{P}_1, v^{P}_2$ and their associated part variables $x_{v^{P}_1}, x_{v^{P}_2}$, representing possible presence of at most $2$ lines. We specify the interaction graph in \textbf{Fig.~\ref{fig:mam}[lower left]}, and the complete factor graph in \textbf{Fig.~\ref{fig:factor_graph}}.  This MAM contains $16$ top-down attention variables $a_{\left\{  v_{\ell}^{P}, v_{i j} \right\} },\ell=1, 2,i=1,\ldots,4,j = 1, 2$, and $10$ lateral attention variables  $a_{\left\{ v_{i_1 1}, v_{i_2 2} \right\} },1 \le i_1, i_2 \le 4,\left| i_1 - i_2 \right| \le 1$. We use the HOF $f_{x_{v^{P}_{\ell}}}, \ell=1, 2$ (\textbf{Fig.~\ref{fig:mam}[upper right]}) to capture the possible presence of line $x_{v^{P}_{\ell}}$ and its top-down interaction with exactly $2$ line segments by
\[
	a_{\sim x_{v^{P}_{\ell}}}^{(k)} = \left\{ a_{\left\{ v^{P}_{\ell}, v_{i k} \right\} }, i=1, \ldots, 4 \right\} ~ \forall k=1, 2;~ ~ \mathcal{B}_{x^{P}_{\ell}} = \left\{ \begin{pmatrix} 0, 0 \end{pmatrix}, \begin{pmatrix} 1, 1 \end{pmatrix}   \right\}\text{; } \mathcal{U}_{x^{P}_{\ell}}(\B{b}) = 0, ~\forall \B{b} \in  \mathcal{B}_{x^{P}_{\ell}}
\] 
This guarantees that a line, if present, has to interact with exactly one line segment in each column.
$\forall i=1,\ldots, 4, ~ j=1, 2$, we use the HOF $f_{x_{v_{i j}}}$ (\textbf{Fig.~\ref{fig:mam}[lower right]}) to capture the possible presence of line segment $v_{i j}$, its top-down interaction with one line and its lateral interaction with one of its continuation candidates by specifying $\mathcal{B}_{x_{v_{i j}}} = \left\{ \begin{pmatrix} 0, 0 \end{pmatrix} , \begin{pmatrix} 1, 1 \end{pmatrix}  \right\}, ~~ \mathcal{U}_{x_{v_{i j}}}(\B{b}) = 0, \forall \B{b} \in \mathcal{B}_{x_{v_{i j}}}$ and
\[
	a_{\sim x_{v_{i j}}}^{(1)} = \left\{ a_{\left\{ v^{P}_{\ell}, v_{i j} \right\} }, ~\ell=1, 2 \right\} ; ~~~ a_{\sim x_{v_{i j}}}^{(2)} = \left\{ a_{\left\{ v_{i j}, v_{\tilde{i} \tilde{j}} \right\} }, ~~ \left| \tilde{i} - i \right| \le 1, ~~ \tilde{j} = 3 - j  \right\} 
\]
The above MAM specifies a uniform distribution on all configurations with at most 2 lines, and coordinately models top-down and lateral interactions between part variables. Top-down interactions model the composition of a line with 2 line segments, and inform lateral interactions to enforce the global constraint of having at most 2 lines. However, they cannot prevent broken lines. Lateral interactions resolve ambiguities in top-down interactions, and ensure proper line segments continuations, which can be read off by looking at the attention variables, even if the line segments themselves are ambiguous about what the continuations are. 

In contrast, an MRF operating only on binary part variables cannot do better than explicitly encoding all valid joint configurations in a single HOF. This is because of a complete lack of conditional independence relationships in the absence of attention variables. As an example, consider the pair of variables $v_{1 2}$ and $v_{4 1}$: if $v_{1 1}, v_{2 2}, v_{3 1}, v_{4 2}$ are \texttt{OFF}, and $v_{2 1}, v_{3 2}$ are \texttt{ON}, then $v_{1 2}$ and $v_{4 1}$ have to be both \texttt{ON} or both \texttt{OFF}. This implies $v_{1 2}$ and $v_{4 1}$ are not conditionally independent given all other variables. In fact, we can use the same argument to establish that, if we do not introduce attention variables, we do not have any pair of part variables that are conditionally independent given the rest. 

\section{Symbol grounding with the sparsifier}

\subsection{Formal definition for the sparsifier}

A sparsifier is a binary higher-order MRF containing two different types of variables: (1) binary part variables, which are the same as the part variables in MAMs and represent the presence/absence of object parts, and (2) binary pixel variables, which represent pixels in a binary image. In this context, we describe the object part associated with a part variable by a set of pixels on the image, and consider the object part as a \textit{parent} of its covered pixels. There is one \texttt{OR} factor per pixel variable, which is a HOF that connects the pixel variable and all its parent part variables.

Formally, for a given $M\times N$ binary image, we have $M N$ pixel variables $I_{i j}, ~1\le i\le M, ~1\le j\le N$ and a set of part variables $\mathcal{V}_S$ in the sparsifier. For a given part variable $x\in \mathcal{V}_S$, we use  $I(x) \subset  \left\{ (i, j): 1\le i \le M, 1\le j\le N \right\} $ to represent its associated object part. Let $x_{\sim I_{i j}} = \left\{ x \in \mathcal{V}_S: \left( i, j \right) \in I(x) \right\} $ be the set of parent part variables for $I_{i j}$. The \texttt{OR} factor $f_{I_{i j}}$ associated with $I_{i j}$ operates on $\left| x_{\sim I_{i j}} \right| + 1$ variables $\left\{ I_{i j} \right\} \cup x_{\sim I_{i j}}$. We represent a configuration for $f_{I_{i j}}$ by its set of entries that are \texttt{ON}, which is an element of $\mathcal{P}\left( \left\{ I_{i j} \right\} \cup x_{\sim I_{i j}} \right)$. Using $f_{I_{i j}}\left( \B{c} \right) $ to denote the potential in log domain of configuration $\B{c} \in \mathcal{P}\left( \left\{ I_{i j} \right\} \cup x_{\sim I_{i j}} \right)$, the \texttt{OR} factor $f_{I_{i j}}$ is defined as
\[
	f_{I_{i j}}(\B{c}) = \left\{
    \begin{array}{lll}
		0, & \text{if } \left| x_{\sim I_{i j}} \cap \B{c} \right| > 0\text{ and } I_{i j} \in \B{c} & \text{(at least one parent  \texttt{ON}, pixel \texttt{ON})}\\
		0, & \text{if } \left| \B{c} \right| = 0 & \text{(all parents  \texttt{OFF}, pixel \texttt{OFF})}\\
		\log\pi_{0 1}, & \text{if } \left| x_{\sim I_{i j}} \cap \B{c} \right| > 0\text{ and } I_{i j} \not\in  \B{c}& \text{(at least one parent  \texttt{ON}, pixel \texttt{OFF})}\\
		\log\pi_{1 0}, & \text{if } \left| x_{\sim I_{i j}} \cap \B{c} \right| = 0\text{ and } I_{i j} \in \B{c} & \text{(all parents \texttt{OFF}, pixel \texttt{ON})}\\
    \end{array}
    \right.
\] 
where $0\le \pi_{0 1}, \pi_{1 0} < 0.5$ and $\log 0$ is assumed to represent $-\infty$. 
\subsection{Illustrative example of sparsifying a line with the sparsifier}\label{sec:sparsifier}

\begin{figure}[t!]
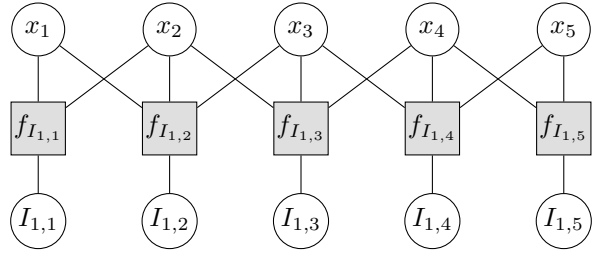

\centering
\tikz{ 
    \node[latent] (i1) {$I_{1,1}$}; 
    \node[latent, right=1 of i1] (i2) {$I_{1,2}$}; 
    \node[latent, right=1 of i2] (i3) {$I_{1,3}$}; 
    \node[latent, right=1 of i3] (i4) {$I_{1,4}$}; 
    \node[latent, right=1 of i4] (i5) {$I_{1,5}$}; 
    \node[obs, rectangle, above=.5 of i1] (f1) {$f_{I_{1,1}}$}; 
    \node[obs, rectangle, above=.5 of i2] (f2) {$f_{I_{1,2}}$}; 
    \node[obs, rectangle, above=.5 of i3] (f3) {$f_{I_{1,3}}$}; 
    \node[obs, rectangle, above=.5 of i4] (f4) {$f_{I_{1,4}}$}; 
    \node[obs, rectangle, above=.5 of i5] (f5) {$f_{I_{1,5}}$}; 
    \node[latent, above=.6 of f1] (p1) {$x_1$}; 
    \node[latent, above=.6 of f2] (p2) {$x_2$}; 
    \node[latent, above=.6 of f3] (p3) {$x_3$}; 
    \node[latent, above=.6 of f4] (p4) {$x_4$}; 
    \node[latent, above=.6 of f5] (p5) {$x_5$}; 
    \factoredge[-] {p1,p2} {f1} {i1};
    \factoredge[-] {p1,p2,p3} {f2} {i2};
    \factoredge[-] {p2,p3,p4} {f3} {i3};
    \factoredge[-] {p3,p4,p5} {f4} {i4};
    \factoredge[-] {p4,p5} {f5} {i5};
}
\caption{Factor graph for the toy example of sparsifying a line with the sparsifier in Sec.~\ref{sec:sparsifier}, with $K=2$.  The \texttt{OR} factor $f_{I_{1 j}}$ operates on the variables $\left\{  I_{1j} \right\} \cup \{x_k: ~ 1\le k \le 3K - 1 \text{ and } |j-k|\le 1 \}$. Each part variable $x_j$ is associated with a unary potential that encourages it to be \texttt{OFF}.}
\label{fig:sparsifier}
\end{figure}

We present herein a simple example to illustrate the \textit{explaining away} type of reasoning implemented by the sparsifier, and discuss how it can be used to sparsify a binary image. Consider a $1\times (3K - 1)$ binary image where all pixels are \texttt{ON}, i.e. a line of length $3K - 1$. We introduce a set of part variables $x_1, \ldots, x_{3K-1}$, and define their corresponding object parts to be $I(x_j) = \left\{ (1, k): 1\le k\le 3K - 1\text{ and }\left| j - k \right| \le 1 \right\}$, i.e. short line segments consisting of two or three pixels. This implies that $x_{\sim I_{1 j}} = \{x_k: ~ 1\le k \le 3K - 1 \text{ and } |j-k|\le 1 \}, ~ 1 \le j \le 3K - 1$, which are sets of size two or three. The \texttt{OR} factor $f_{I_{1 j}}$ associated with $I_{1 j}$ operates on the $\left| x_{\sim I_{1 j}} \right| + 1$ variables $\left\{ I_{1 j} \right\} \cup x_{\sim I_{j}}$.
See \textbf{Fig. \ref{fig:sparsifier}} for the complete factor graph in the case $K=2$.

We associate a unary potential in the log domain of $0$ for being \texttt{OFF} and  $\log\pi$ for being \texttt{ON} with each part variable, and set $0 < \pi_{10} < \pi < 1$.
Under this setup, all the optimal joint configurations have exactly $K$ part variables being \texttt{ON}, and each such joint configuration corresponds to a sparse representation of the line. 
In the case $K = 2$ shown in \textbf{Fig.~\ref{fig:sparsifier}}, these optimal joint configurations are $\{x_1, x_4\}$, $\{x_2, x_4\}$ and $\{x_2, x_5\}$:
\begin{enumerate}
    \item If all the part variables are \texttt{OFF}, the potential of the joint configuration is $5 \log \pi_{10}$.
    \item If one part variable is \texttt{ON}, the potential of the joint configuration is $\log \pi + 2 \log \pi_{10}$ (when $x_2$, $x_3$, or $x_4$ is \texttt{ON}) or $\log \pi + 3 \log \pi_{10}$ (when $x_1$ or $x_5$ is \texttt{ON}),
    \item The three joint configurations $\{x_1, x_4\}$, $\{x_2, x_4\}$ and $\{x_2, x_5\}$ all have potential $2 \log \pi$.
    \item If more than three part variables are \texttt{ON} the joint potential is at most $3 \log \pi$.
\end{enumerate}
Since $0 < \pi_{10} < \pi < 1$, we establish the optimality of $\{x_1, x_4\}$, $\{x_2, x_4\}$ and $\{x_2, x_5\}$.

\subsection{Learning object parts from binary images}

We use a special case of hierarchical compositional network (HCN)~\cite{lazaro2016hierarchical}, where we have a single layer of latent variables and no \texttt{POOL} factors, to learn object parts directly from binary images. Intuitively, this is equivalent to replacing the predefined object parts by a set of binary weight variables covering a regular region (e.g. a rectangular region) in the sparsifier, and do MPBP inference jointly on the part, pixel and weight variables. Using the inference results from MPBP, we decode a set of \texttt{ON} weight variables which we use to define our learned object parts. The weight variables are shared convolutionally and across multiple images, allowing us to aggregate information from all images in the learning process and learn a small number of meaningful, generic and reusable object parts.

\section{MPBP updates for involved factors} \label{sec:map_updates}

Due to the simple structures in the HOFs in MAMs and the \texttt{OR} factors in the sparsifier, MPBP updates for these factors can be done efficiently.

Use $\mathbb{M}^{u\to v}\in \mathbb{R}^{2}$ to denote the message (in log domain) from $u$ to $v$, where $u, v$ correspond to a pair of factor and variable. Since we can arbitrarily normalize the messages, for convenience and efficiency, in our implementations we use $m_{u\to v} = \mathbb{M}^{u\to v}_2 - \mathbb{M}^{u\to v}_1$ as our messages (i.e. we normalize the messages by substracting $\mathbb{M}^{u\to v}_1$ so that $\mathbb{M}^{u\to v}_1$ is always assumed to be 0).

\subsection{MPBP updates for the HOFs in MAMs: } 

Due to the simple structure of the HOFs in MAMs, MPBP updates can be done efficiently. For the HOF $f_{x_v}$ in a MAM, given the messages $m_{x_v\to f_{x_v}}, ~ m_{a \to f_{x_v}}, ~ a \in a_{\sim x_v}$ from variables to $f_{x_v}$, $\forall k=1, \ldots, M_{x_v}$, define
\begin{align*}
	m^{(k)}_1 &= \max_{a \in a_{\sim x_v}^{(k)}} m_{a \to f_{x_v}},  ~~~ a^{(k)} = \argmax_{a \in a_{\sim x_v}^{(k)}} m_{a \to f_{x_v}}\\
	m^{(k)}_2 &= \begin{cases}
		\max \limits_{a \in a_{\sim x_v}^{(k)}:  ~ a\neq  a^{(k)}}m_{a \to f_{x_v}} &\text{if }| a^{(k)}_{\sim x_v} | > 1\\ 
		0 & \text{if } | a^{(k)}_{\sim x_v} | = 1,
\end{cases}
\end{align*}
so that $m^{(k)}_1$ denotes the largest message coming from all the elements in the set $a^{(k)}$ and $m^{(k)}_2$ denote the second largest message.

For a pair of factor $f$ and variable $v$, the MPBP updates $m_{f\to v}$ are given by $\mathbb{M}^{f \to v}_2 - \mathbb{M}^{f\to v}_1$, where
\begin{align*}
	\mathbb{M}^{f_{x_v}\to x_v}_2 & = \max_{b \in \mathcal{B}_{x_v}, \| \B{b} \|_1 > 0} \left[\mathcal{U}_{x_v}(\B{b}) + m_{x_v \to f_{x_v}} + \sum_{k=1}^{M_{x_v}} b_k m^{(k)}_1 \right]\\
	\mathbb{M}^{f_{x_v}\to x_v}_1 &=  \mathcal{U}_{x_v}(\begin{pmatrix} 0, \ldots, 0 \end{pmatrix} )\\
 	\mathbb{M}^{f_{x_v}\to a}_2 & =  \max_{\B{b} \in \mathcal{B}_{x_v}: ~ b_k = 1} \left[\mathcal{U}_{x_v}(\B{b}) + m_{x_v \to f_{x_v}} + \sum_{\ell\neq k}b_{\ell} m^{(\ell)}_1 \right], ~ a \in a_{\sim x_v}^{(k)}, ~ a \neq a^{(k)}\\
	\mathbb{M}^{f_{x_v}\to a}_1 & =  \max_{\B{b} \in \mathcal{B}_{x_v}} \left[ \mathcal{U}_{x_v}(b) + \mathbbm{1}_{\| \B{b} \|_1 > 0}m_{x_v\to f_{x_v}} + \sum_{\ell=1}^{M_{x_v}} b_{\ell} m^{(\ell)}_1 \right], ~ a \in a_{\sim x_v}^{(k)}, ~ a \neq a^{(k)}\\
	\mathbb{M}^{f_{x_v}\to a^{(k)}}_2 & = \max_{\B{b} \in \mathcal{B}_{x_v}: ~ b_k = 1} \left[\mathcal{U}_{x_v}(\B{b}) + m_{x_v \to f_{x_v}} + \sum_{\ell\neq k}b_{\ell} m^{(\ell)}_1 \right] \\
	\mathbb{M}^{f_{x_v}\to a^{(k)}}_1 & = \max_{\B{b} \in \mathcal{B}_{x_v}}\left[ \mathcal{U}_{x_v}(\B{b}) + \mathbbm{1}_{\|\B{b}\|_1 > 0}m_{x_v\to f_{x_v}}  + \sum_{\ell\neq k} b_{\ell} m^{(\ell)}_1 + b_k m^{(k)}_2\right] \\
\end{align*}

\subsection{MPBP updates for the \texttt{OR} factors in the sparsifier: } 

For a \texttt{OR} factor $f_{I_{i j}}$ in the sparsifier, if $\left| x_{\sim I_{i j}} \right| = 1$, the updates are trivial. If $\left| x_{\sim I_{i j}} \right| > 1$, given the messages $m_{I_{i j}\to f_{I_{i j}}}, m_{x \to f_{I_{i j}}}, x \in x_{\sim I_{i j}}$, define $x^{(i j)}_1 = \argmax_{x \in  x_{\sim I_{i j}}} m_{x \to f_{I_{i j}}}, x^{(i j)}_2 = \argmax_{x \in x_{\sim I_{i j}}, x \neq x^{(i j)}_1} m_{x\to f_{I_{i j}}}$. 

Use $m^{+}$ to denote $\max\left\{ 0, m \right\}$, and $m \vee n$ to denote $\max\left\{m, n\right\}$. For a pair of factor $f$ and variable $v$, the MPBP updates $m_{f\to v}$ are given by $\mathbb{M}^{f \to v}_2 - \mathbb{M}^{f\to v}_1$, where
\begin{align*}
	\mathbb{M}^{f_{I_{i j}} \to I_{i j}}_2 & =  \left[ m_{x^{(i j)}\to f_{I_{i j}}} + \sum_{x \neq x^{(i j)}} m_{x\to f_{I_{i j}}}^{+} \right]  \vee \log\pi_{1 0} \\
	\mathbb{M}^{f_{I_{i j}} \to I_{i j}}_1 &= \left( \log\pi_{0 1} +  m_{x^{(i j)}\to f_{I_{i j}}} + \sum_{x \neq x^{(i j)}} m_{x\to f_{I_{i j}}}^{+}  \right)^{+}\\
	\mathbb{M}^{f_{I_{i j}} \to x^{(i j)}_1}_2 & =  m_{I_{i j} \to  f_{I_{i j}}} \vee \log\pi_{0 1} + \sum_{q \in x_{\sim I_{i j}}, ~ q \neq  x^{(i j)}_1} m_{q \to f_{I_{i j}}}^{+} \\
	\mathbb{M}^{f_{I_{i j}} \to x^{(i j)}_1}_1 & = \left( m_{I_{i j}\to f_{I_{i j}}} + \log\pi_{1 0} \right)^{+} \vee \left(  m_{I_{i j} \to f_{I_{i j}}} \vee \log\pi_{0 1}  + m_{x^{(i j)}_2\to f_{I_{i j}}} + \sum_{q \neq  x, x^{(i j)}_2} m_{q \to f_{I_{i j}}}^{+} \right) \\
\end{align*}
and $\forall x \neq x^{(i j)}_1$,
\begin{align*}
	\mathbb{M}^{f_{I_{i j}} \to x}_2 &=  m_{I_{i j} \to  f_{I_{i j}}} \vee \log\pi_{0 1} + \sum_{q \in x_{\sim I_{i j}}, ~ q \neq  x} m_{q \to f_{I_{i j}}}^{+}, ~ \forall x \neq x^{(i j)}_1\\
	\mathbb{M}^{f_{I_{i j}} \to x}_1 &= \left( m_{I_{i j}\to f_{I_{i j}}} + \log\pi_{1 0} \right)^{+} \vee \left(m_{I_{i j} \to f_{I_{i j}}} \vee \log\pi_{0 1} + m_{x^{(i j)}_1\to f_{I_{i j}}} + \sum_{q \in x_{\sim I_{i j}}, ~ q \neq  x, x^{(i j)}_1} m_{q \to f_{I_{i j}}}^{+}\right)\\
\end{align*}

\section{Hardware used for experiments}

We use machines with 32 Intel Xeon E5-2620 CPUs and 2 GeForce RTX 2080 Ti GPUs for experiments.

\section{Details for the pathfinder challenge}
\subsection{Dataset details}

\begin{figure}[h!]
	\centering
	\includegraphics[width=\textwidth]{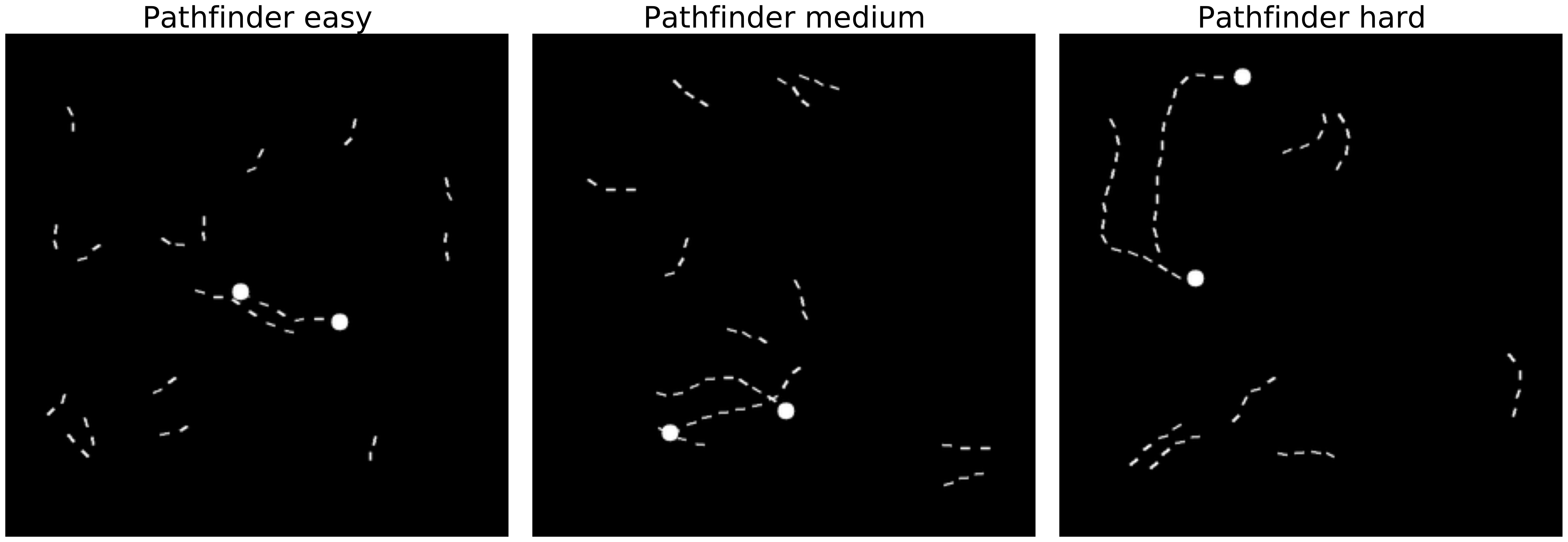}
	\caption{Pathfinder example images from the 3 difficulty levels (\textit{easy, medium} and \textit{hard}).}
	\label{fig:pathfinder_example}
\end{figure}

The \textit{easy, medium} and \textit{hard} pathfinder datasets used in our experiments were generated using code provided by the authors, available at \url{https://github.com/drewlinsley/pathfinder}. See \textbf{Fig.~\ref{fig:pathfinder_example}} for example images from the 3 difficulty levels.

The code was released under the MIT license.

\subsection{Model}

The MAM for the pathfinder challenges uses $T\times M \times N$ part variables $x_{t r c}, 1\le t\le T, 1\le r\le M, 1\le c\le N$ to represent the possible presence of $T$ types of object parts in an $M \times N$ grid of locations.  Let $\left\{ (t_1, t_2, \Delta r, \Delta c), ~ (t_2, t_1, -\Delta r, -\Delta c)\right\} , ~ 1 \le t_1, t_2 \le T, ~\Delta r, ~\Delta c \in \mathbb{Z}$ denote the co-occurrence of $2$ object parts of types $t_1, t_2$, where the relative displacement of the type $t_2$ object part with respect to the type $t_1$ object part is $(\Delta r, \Delta c)$. Each co-occurrence $\left\{ (t_1, t_2, \Delta r, \Delta c), ~ (t_2, t_1, -\Delta r, -\Delta c)\right\}$ is applied convolutionally and defines a set of attention variables $\left\{ a_{\left\{ x_{t_1 r_1 c_1}, x_{t_2 r_2 c_2} \right\} }: r_2 - r_1 = \Delta r, c_2 - c_1 = \Delta c, 1\le r_1, r_2 \le  M, 1 \le c_1, c_2 \le N \right\} $. The attention variables in the MAM are given by the union of the above sets of attention variables. 
We learn $T=16$ types of object parts (\textbf{Fig.~\ref{fig:pathfinder_sparsification}[left]}) from 10 training images, and extract co-occurences of object parts from data. For $x_{t r c}$, we partition its associated attention variables $a_{\sim x_{t r c}}$ into $2$ sets to capture possible contour continuations on both sides. The HOF $f_{x_{t r c}}$ is specified by (1) the set of prototypical interaction patterns $\mathcal{B}_{x_{t r c}} = \left\{ \begin{pmatrix} 0, 0 \end{pmatrix} , \begin{pmatrix} 0, 1 \end{pmatrix} , \begin{pmatrix} 1, 0 \end{pmatrix} , \begin{pmatrix} 1, 1 \end{pmatrix}  \right\} $, which indicates that a contour should continue on one or both sides, and (2) the potential in log domain $\mathcal{U}_{x_{t r c}}\left( \begin{pmatrix} 0, 0 \end{pmatrix}  \right) = \mathcal{U}_{x_{t r c}}\left( \begin{pmatrix} 1, 1 \end{pmatrix}  \right) = 0, \mathcal{U}_{x_{t r c}}\left( \begin{pmatrix} 0, 1 \end{pmatrix}  \right) = \mathcal{U}_{x_{t r c}}\left( \begin{pmatrix} 1, 0 \end{pmatrix}  \right) = -1.6 $, which penalizes contour termination and encourages the contour to be \texttt{OFF} or continue on both sides. We connect the MAM to the image with the sparsifier to model the presence of a variable number of broken contours.

\subsection{Pathfinder sparsification example}

\begin{figure}[h!]
	\centering
	\includegraphics[width=\textwidth]{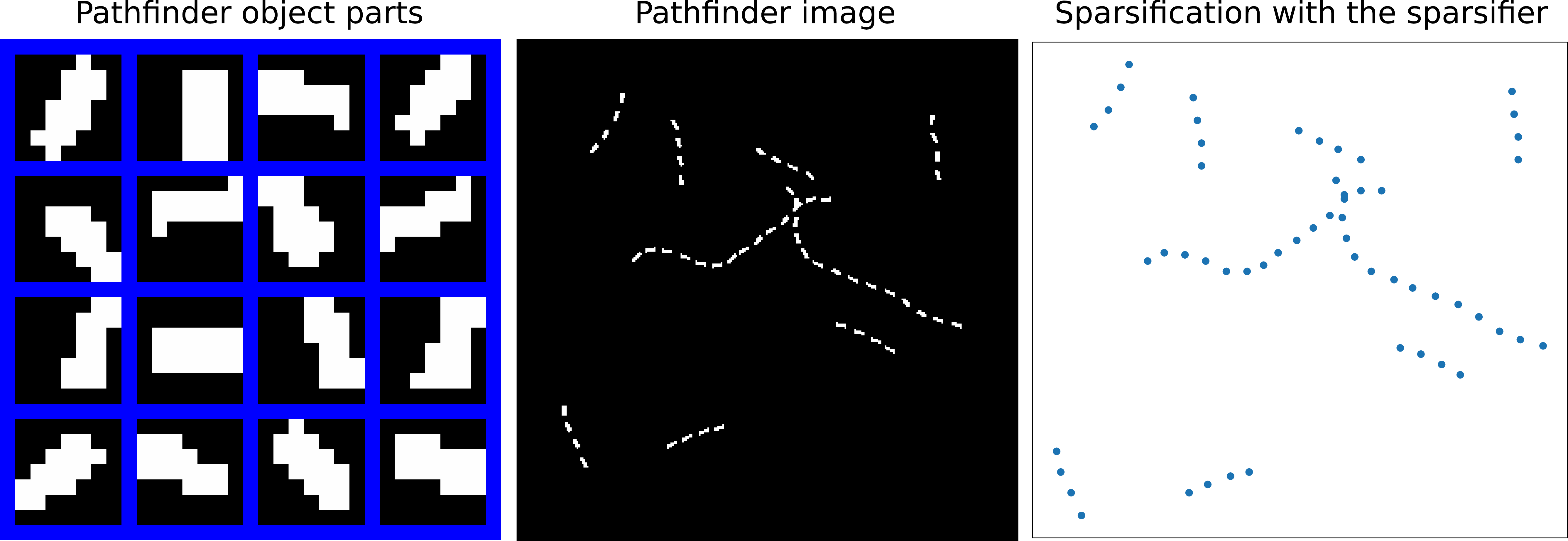}
	\caption{Sparsifying pathfinder images.}
	\label{fig:pathfinder_sparsification}
\end{figure}

See \textbf{Fig.~\ref{fig:pathfinder_sparsification}} for an example of sparsifying pathfinder images.

\subsection{Learning}

\textbf{Learning: }
Learning of object parts and sparsifications are done with $\pi_{0 1} = 0.07, \pi_{1 0} = 0.0019$, and a unary potential in log domain of $0$ for \texttt{OFF} and $-20$ for \texttt{ON} for all part variables to promote sparse object part activations. For sparsifications, we repeat MPBP 10 times for each individual contour to account for randomness. We define the MAM by aggregrating co-occurences from all individual contours while filtering out any co-occurence whose relative frequency is less than $1.7\times 10^{-7}$.

\subsection{Pathfinder ultimate}

In the original pathfinder datasets, for \textit{easy}, there are two long contours of length 6, and 15 distractor contours of length 2 (\textbf{Fig.~\ref{fig:pathfinder_example}[left]}). For \textit{medium}, there are two long contours of length 9, and 9 distractor contours of length 3 (\textbf{Fig.~\ref{fig:pathfinder_example}[center]}). For \textit{hard}, there are two long contours of length 14, and 7 distractor contours of length 4 (\textbf{Fig.~\ref{fig:pathfinder_example}[right]}).

Since only the \textit{hard} difficulty level proves to be challenging, we generate a new \textit{ultimate} difficulty level by modifying the original data generation code to further probe the generalization capacities of MAM and H-CNN. For each \textit{ultimate} image: (1) we first draw two long contours, each one with a length selected uniformly in $\left[20, 30\right]$; (2) we then draw two long distractors, each one with a length selected uniformly in $\left[15, 30\right]$ (3) we finally draw at most $8$ short distractors with length selected uniformly in $\left[6, 14\right]$. See \textbf{Fig.~4[center right]} in the main text for an example.
\section{Details for the cABC challenge}
\subsection{Dataset details}

We use datasets provided by the authors at \url{https://openreview.net/forum?id=HJxrVA4FDS}, available for download at \url{https://bit.ly/2wdQYGd}. See \textbf{Fig.~\ref{fig:pathfinder_example}} for example images from the 3 difficulty levels.

\begin{figure}[h!]
	\centering
	\includegraphics[width=\textwidth]{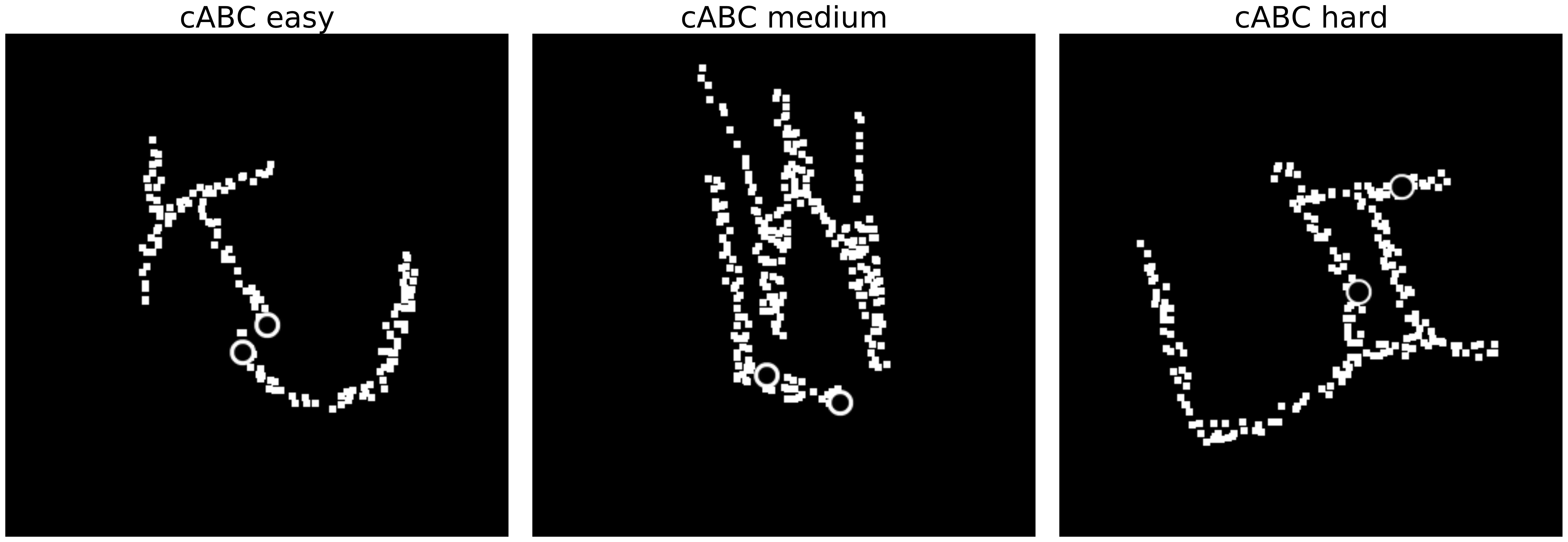}
	\caption{cABC example images from the 3 difficulty levels (\textit{easy, medium} and \textit{hard}).}
	\label{fig:cabc_example}
\end{figure}

\subsection{Model}

We model \textit{elastic graphs} with object-specific MAMs, and match elastic graphs to noisy, distorted letters to solve the cABC challenge. An elastic graph $\mathcal{G}_E(\eta) =\left( \mathcal{V}_E, \mathcal{E}_E, \eta \right) $ is an undirected graph where $\mathcal{V}_E$ represents its set of vertices and $\mathcal{E}_E = \left\{ \left\{ u, v \right\}: ~ u, v\in \mathcal{V}_E, ~ \exists \text{ an edge in }\mathcal{G}_E \text{ connecting }u, v  \right\} $. Each $v \in  \mathcal{V}_E$ represents an object part, and each $\left\{ u, v \right\} \in \mathcal{E}_E$ represents an \textit{elastic constraint} on $u, v$. An elastic graph models the letter as a composition of all the object parts represented by the vertices. We associate a reference location $(r_v, c_v)$ with $v \in \mathcal{V}_E$, which indicates the expected location of the corresponding object part. We allow the vertex $v\in \mathcal{V}_E$ to move around, and use $\left( \tilde{r}_v, \tilde{c}_v \right) $ to denote its actual location. We use $\eta \ge 1$ to restrict the actual locations of the vertices to be $0\le \tilde{r}_v - r_v, \tilde{c}_v - c_v < \eta, \forall v \in \mathcal{V}_E$. We associate a \textit{perturb radius} $\gamma_{\left\{ u, v \right\}} $ with each edge $\left\{ u, v \right\}  \in \mathcal{E}_E$, which constrains the actual relative displacements $\left(\tilde{r}_{v} - \tilde{r}_{u}, \tilde{c}_{v} - \tilde{c}_{u}\right)$ to differ (in $L_{\infty}$ norm) by no more than $\gamma_{\left\{ u, v \right\} }$ from the reference relative displacements $\left( r_{v} - r_{u}, c_{v} - c_{u} \right) $, i.e.  $ \left| (\tilde{r}_{v} - \tilde{r}_{u}) - (r_{v} - r_{u}) \right| \le \gamma_{\left\{ u, v \right\} }, \left| (\tilde{c}_{v} - \tilde{c}_{u}) - (c_{v} - c_{u}) \right| \le \gamma_{\left\{ u, v \right\} } $.

\subsection{Formal definitions of the object-specific MAM}

Elastic graphs are inherently shift invariant, and are completely specified by the relative displacements of its vertices. For ease of presentation, we define $\left( r^{\mathcal{G}_E(\eta)}, c^{\mathcal{G}_E(\eta)} \right) = \left( \min_{v \in \mathcal{V}_E}r_{v}, \min_{v \in \mathcal{V}_E} c_{v} \right)$ as the \textit{anchor point} of $\mathcal{G}_E(\eta)$ with reference locations $\left\{ \left( r_v, c_v \right) , v \in \mathcal{V}_E \right\} $. We can transform an elastic graph $\mathcal{G}_E(\eta)$ with anchor point $\left( r^{\mathcal{G}_E(\eta)}, c^{\mathcal{G}_E(\eta)} \right)$ into an elastic graph $\overline{\mathcal{G}}_E(\eta)$ with anchor point $\left( r^{\overline{\mathcal{G}}_E(\eta)}, c^{\overline{\mathcal{G}}_E(\eta)} \right) $ by updating the reference location of vertex $v\in \mathcal{V}_E$ from $\left( r_v, c_v \right) $ to $\left( r_v - r^{\mathcal{G}_E(\eta)} + r^{\overline{\mathcal{G}}_E(\eta)}, c_v - c^{\mathcal{G}_E(\eta)} + c^{\overline{\mathcal{G}}_E(\eta)} \right) $.

We can specify an object-specific MAM for a given elastic graph $\mathcal{G}_E(\eta)$. We associate a part variable $x_{v \tilde{r}_v \tilde{c}_v}^{\mathcal{G}_E(\eta)}$ with each allowed location $\left( \tilde{r}_v, \tilde{c}_v \right), v \in \mathcal{V}_E, 0\le \tilde{r}_v - r_v, \tilde{c}_v - c_v < \eta $. For each edge $\left\{ u, v \right\} \in \mathcal{E}_E$, we associate a lateral attention variable $a_{\left\{ x^{\mathcal{G}_E(\eta)}_{u \tilde{r}_{u} \tilde{c}_{u}}, x^{\mathcal{G}_E(\eta)}_{v \tilde{r}_{v} \tilde{c}_{v}} \right\} }$ with each pair of locations $\left( \tilde{r}_u, \tilde{c}_u \right), \left( \tilde{r}_v, \tilde{c}_v \right) $ in an allowed relative displacement (i.e. $\left| (\tilde{r}_{v} - \tilde{r}_{u}) - (r_{v} - r_{u}) \right| \le \gamma_{\left\{ u, v \right\} }, \left| (\tilde{c}_{v} - \tilde{c}_{u}) - (c_{v} - c_{u}) \right| \le \gamma_{\left\{ u, v \right\} } $). We additionally introduce the part variable $x^{\mathcal{G}_E(\eta)}$ to represent the possible presence of the letter, and the top-down attention variables
\[a_{\left\{ x^{\mathcal{G}_E(\eta)}, x^{\mathcal{G}_E(\eta)}_{v \tilde{r}_v \tilde{c}_v} \right\} }, v \in \mathcal{V}_E, 0\le \tilde{r}_v - r_v, \tilde{c} - c_v < \eta\]
For the HOF $f_{x^{\mathcal{G}_E(\eta)}}$, we partition $a_{\sim x^{\mathcal{G}_E(\eta)}}$ into $M_{x^{\mathcal{G}_E(\eta)}} = \left| \mathcal{V}_E \right| $ disjoint sets of attention variables
\[ 
	a_{\sim x^{\mathcal{G}_E(\eta)}}^{(k)} = \left\{ a_{\left\{ x^{\mathcal{G}_E(\eta)}, x^{\mathcal{G}_E(\eta)}_{v_k \tilde{r}_{v_k}, \tilde{c}_{v_k}} \right\} }:  0\le \tilde{r}_{v_k} - r_{v_k}, \tilde{c}_{v_k} - c_{v_k} < \eta\right\}, k=1, \ldots, \left| \mathcal{V}_E \right| 
\]
where $\mathcal{V}_E = \left\{ v_1, \ldots, v_{\left| \mathcal{V}_E \right| } \right\} $ is some ordering of all the vertices in $ \mathcal{V}_E$. Define $\mathcal{B}_{x^{\mathcal{G}_E(\eta)}} = \left\{ \begin{pmatrix} 0, \ldots, 0 \end{pmatrix}, \begin{pmatrix} 1, \ldots, 1 \end{pmatrix}   \right\} $ and $\mathcal{U}_{x^{\mathcal{G}_E(\eta)}}(b) = 0, \forall b \in \mathcal{B}_{x^{\mathcal{G}_E(\eta)}}$. Intuitively, this enforces that, if the letter is present, then each $v\in \mathcal{V}_E$ much be present, at exactly one of the allowed location.

For the HOFs $f_{x^{\mathcal{G}_E(\eta)}_{v \tilde{r}_v \tilde{c}_v}}, v\in \mathcal{V}_E, 0\le \tilde{r}_v - r_v, \tilde{c}_v - c_v < \eta$, we partition $a_{\sim x^{\mathcal{G}_E(\eta)}_{v \tilde{r}_v \tilde{c}_v}}$ into $M^{\mathcal{G}_E(\eta)}_v = \left| \left\{ u \in \mathcal{V}_E: \left\{ u, v \right\} \in \mathcal{E}_E \right\}  \right|  + 1$ disjoint sets of attention variables
\[
	a_{\sim x^{\mathcal{G}_E(\eta)}_{v \tilde{r}_v \tilde{c}_v}}^{(k)} = a_{\sim x^{\mathcal{G}_E(\eta)}_{v \tilde{r}_v \tilde{c}_v}} \cap a_{\sim x^{\mathcal{G}_E(\eta)}_{u_{k} \tilde{r}_{u_{k}} \tilde{c}_{u_{k}}}}, k=1, \ldots, M^{\mathcal{G}_E(\eta)}_v - 1, a_{\sim x^{\mathcal{G}_E(\eta)}_{v \tilde{r}_v \tilde{c}_v}}^{\left(M^{\mathcal{G}_E(\eta)}_v\right) } = \left\{ a_{x^{\mathcal{G}_E(\eta)}, x^{\mathcal{G}_E(\eta)}_{v \tilde{r}_v \tilde{c}_v}} \right\}
\]
where we impose some ordering $\left\{ u\in \mathcal{V}_E: \left\{ u, v \right\} \in \mathcal{E}_E \right\}  =  \left\{ u_1, \ldots, u_{M_{x^{\mathcal{G}_E(\eta)}_{v \tilde{r}_v \tilde{c}_v}} - 1} \right\} $. We define $\mathcal{B}_{x^{\mathcal{G}_E(\eta)}_{v \tilde{r}_v \tilde{c}_v}} = \left\{ \begin{pmatrix} 0, \ldots, 0 \end{pmatrix} , \begin{pmatrix} 1, \ldots, 1 \end{pmatrix}  \right\} $ and $ \mathcal{U}_{x^{\mathcal{G}_E(\eta)}_{v \tilde{r}_v \tilde{c}_v}}\left( \begin{pmatrix} 0, \ldots, 0 \end{pmatrix}  \right) = -1000, \mathcal{U}_{x^{\mathcal{G}_E(\eta)}_{v \tilde{r}_v \tilde{c}_v}}(\begin{pmatrix} 1, \ldots, 1 \end{pmatrix} ) = 0 $.

For the cABC challenge, assume the sizes of the images imply that object parts can be instantiated on an $M\times M$ grid. We extract elastic graphs from $N$ letters in the training images using the procedure outlined in Sec.~\ref{sec:learn_elastic_graphs}, and transform them into a set of elastic graphs $\mathcal{G}_E^{(n)}(M) =  \left( \mathcal{V}_E^{(n)}, \mathcal{E}_E^{(n)}, M \right), n=1, \ldots, N$ whose anchor points are all $(1, 1)$. We construct $N$ MAMs $\mathcal{M}^{\mathcal{G}_E^{(n)}(M)}, n=1, \ldots, N$ as described above, and merge them into a single MAM $\mathcal{M}$ with part variables that live within the $M\times M$ grid $x^{\mathcal{G}_E^{(n)}(M)}_{v \tilde{r}_v \tilde{c}_v}, v\in \mathcal{V}_E^{(n)}, r_v \le \tilde{r}_v \le  M, c_v \le \tilde{c}_v \le M, n=1, \ldots, N$, the lateral attention variables $a_{\left\{ x^{\mathcal{G}_E^{(n)}(M)}_{u \tilde{r}_{u} \tilde{c}_{u}}, x^{\mathcal{G}_E^{(n)}(M)}_{v \tilde{r}_{v} \tilde{c}_{v}} \right\} }$ where $ \left\{ u, v \right\} \in \mathcal{E}_E^{(n)}, \left| (\tilde{r}_{v} - \tilde{r}_{u}) - (r_{v} - r_{u}) \right| \le \gamma_{\left\{ u, v \right\} }, \left| (\tilde{c}_{v} - \tilde{c}_{u}) - (c_{v} - c_{u}) \right| \le \gamma_{\left\{ u, v \right\} }, r_v \le \tilde{r}_v \le  M, c_v \le \tilde{c}_v \le M, n=1, \ldots, N $ and collapsing the part variables $x^{\mathcal{G}_E^{(n)}(M)}$ into a single part variable $x^{\mathcal{M}}$, with top-down attention variables
\[ a_{\left\{ x^{\mathcal{M}}, x^{\mathcal{G}_E^{(n)}(M)}_{v \tilde{r}_v \tilde{c}_v} \right\} }, r_v \le \tilde{r}_v \le  M, c_v \le \tilde{c}_v \le M, n=1, \ldots, N \]

The HOFs $f_{x^{\mathcal{G}_E^{(n)}(M)}_{v \tilde{r}_v \tilde{c}_v}}, r_v \le \tilde{r}_v \le  M, c_v \le \tilde{c}_v \le M, n=1, \ldots, N$ remain the same (with $x^{\mathcal{G}_E^{(n)}(M)}$ replaced by $x^{\mathcal{M}}$), while for the HOF $f_{x^{\mathcal{M}}}$, we now partition $a_{\sim x^{\mathcal{M}}}$ into $M_{x^{\mathcal{M}}} = \sum_{n=1}^{N} \left| \mathcal{V}_E^{(n)} \right| $ disjoint sets of attention variables: $\forall n=1, \ldots, N, k=1, \ldots, \left| \mathcal{V}_E^{(n)} \right|$,
\[
	a_{\sim x^{\mathcal{M}}}^{(\sum_{i=1}^{n-1} \left| \mathcal{V}_E^{(n)} \right| + k )} = \left\{ a_{\left\{ x^{\mathcal{M}}, x^{\mathcal{G}_E^{(n)}(M)}_{v_k \tilde{r}_{v_k}, \tilde{c}_{v_k}} \right\} }:  r_v \le \tilde{r}_v \le  M, c_v \le \tilde{c}_v \le M\right\}
\]
and $\mathcal{B}_{x^{\mathcal{M}}} = \left\{ \begin{pmatrix} 0, \ldots, 0 \end{pmatrix}  \right\} \cup \left\{ b^{(n)}, n=1, \ldots, N \right\} $, where $\forall n \in \left\{ 1, \ldots, N \right\} $,
\[
b^{(n)}_j = \begin{cases}
1, & \text{if } \sum_{i=1}^{n-1} \left| \mathcal{V}_E^{(i)} \right| < j \le \sum_{i=1}^{n} \left| \mathcal{V}_E^{(i)} \right|\\
0, & \text{otherwise}
\end{cases}
\]
and 
\[
	\mathcal{U}_{x^{\mathcal{M}}}(\begin{pmatrix} 0, \ldots, 0 \end{pmatrix}) = -1000,  \mathcal{U}_{x^{\mathcal{M}}}(b) = 0, \forall b \in \mathcal{B}_{x^{\mathcal{M}}}, b \neq \begin{pmatrix} 0, \ldots, 0 \end{pmatrix} 
\]
Intuitively, each individual $ \mathcal{M}^{\mathcal{G}_E^{(n)}(M)}$ encodes the presence of the $n$th letter at a certain location on the image, while $ \mathcal{M}$ merges the individual MAMs to encode the presence of exactly one letter at a certain location on the image. We connect $2$ copies $\mathcal{M}^{(1)}, \mathcal{M}^{(2)}$ of $\mathcal{M}$ to the image with the sparsifier to model the presence of $2$ noisy, distorted letters in each image.

\subsection{cABC sparsification example}

See \textbf{Fig. 2[center right]} and \textbf{Fig. 2[right]} in the main text for an example of sparsifying a noisy, distorted cABC letter. The vertices in the elastic graph in \textbf{Fig. 2[right]} in the main text represent the sparse set of object part activations given by sparsifying the binary input image (\textbf{Fig. 2[center right]} in the main text) with the sparsifier using a single object part learned from 10 training images (lower left of \textbf{Fig. 2[right]} in the main text).

\subsection{Learning elastic graphs for cABC} \label{sec:learn_elastic_graphs}

Learning of the single object part and sparsifications are done using $\pi_{0 1} = 0.45, \pi_{1 0} = 0.05$, and a unary potential in log domain of $0$ for \texttt{OFF} and $-20$ for \texttt{ON} for all part variables to promote sparse object part activations. We define the vertices in the elastic graph using the sparse object part activations from MPBP decoding.

Given a sparse set of object part activations ($\mathcal{V}_E$), we use \textbf{Alg.~\ref{alg:add_edges}} to construct an elastic graph. In our experiments, we used perturbation factor $7.0$, tolerance $2.0$ and max connection length $200$. These hyper-parameters were derived by doing a  parameter sweep over integers and measuring performance on a small validation set.

Intuitively, we use the perturbation factor to capture the notion that the relative displacement of a pair of far-away object parts can vary more than that of a pair of close-by object parts. For a pair of object parts $u, v \in \mathcal{V}_E$, using $d_{u, v}$ to denote their Euclidean distance, the associated perturb radius is roughly given  by $d_{u, v}$ divided by the perturb factor. To implement the constrained elasticities, a naive approach is to add an elastic constraint for every pair of vertices. However, this would result in an overly dense elastic graph and makes subsequent inference inefficient. In \textbf{Alg.~\ref{alg:add_edges}}, we use a heuristic algorithm to add a small number of elastic constraints. We start from vertices that are closest together, and progressively add elastic constraints. For a given pair of vertices, we skip adding their elastic constraint if the desired elastic constraint can already be enforced using existing elastic constraints (up to the specified tolerance). We adopt a final refining step to account for rounding errors. Refer to \textbf{Alg.~\ref{alg:add_edges}} for more details. See \textbf{Fig. 2[right]} in the main text for an example elastic graph.

\begin{algorithm}[htbp]
\footnotesize
\SetAlgoLined\DontPrintSemicolon
\SetKwFunction{GetEdges}{GetEdges}
\SetKwFunction{GetPotentialEdges}{GetPotentialEdges}
\SetKwFunction{AddEdges}{AddEdges}
\SetKwFunction{EdgeDepthFirstSearch}{EdgeDepthFirstSearch}
\SetKwFunction{RefineEdges}{RefineEdges}

\KwIn{ Elastic graph vertices $\mathcal{V}_E$ and the associated reference locations $(r_v, c_v), \forall v \in \mathcal{V}_E$, perturbation factor $p$, tolerance $t$, max connection length $l$ }
\KwOut{Elastic graph edges $\mathcal{E}_E$ and the associated perturb radius $\gamma_{\left\{ u, v \right\} }, \forall \left\{ u, v \right\} \in \mathcal{E}_E$}

\SetKwProg{myproc}{Procedure}{}{}
\myproc{\GetEdges{$\mathcal{V}_E$, $r$, $c$, $p$, $t$, $l$}}{
    $C, d \gets$ \GetPotentialEdges($\mathcal{V}_E$, $r$, $c$, $l$);

    $\mathcal{E}_E, \gamma \gets$ \AddEdges($\mathcal{C}$, $d$, $p$, $t$);

    $O \gets$ \EdgeDepthFirstSearch{$\mathcal{E}_E$, [], \mbox{\textbf{AnyElementOf}} ($\mathcal{V}_E$) };
    
    $\gamma \gets$ \RefineEdges{$O$, $d$, $\gamma$, $p$};

   \KwRet $\mathcal{E}_E$, $\gamma$;
}
\myproc{\GetPotentialEdges{$\mathcal{V}_E$, $r$, $c$, $l$}}{
    $P \gets []$
    
    \For{$u$ \textbf{in} $\mathcal{V}_E$}{
        \For{$v$ \textbf{in} $\mathcal{V}_E$}{
            $d_{u,v} \gets \sqrt{(r_v - r_u)^2 + (c_v  - c_u)^2}$\\
            \If{$d_{u,v} < l$ \mbox{\textbf{and}} $\left\{ u, v \right\} \notin P$ }{
		    $P \gets $\textbf{append}($P$, $\left\{ u, v \right\} $)
            }
        }
    }
    $C \gets $\textbf{sorted}($P$) \textbf{by} $d_{u,v}$ \textbf{ascending}
    
    \KwRet $C$, $d$;
}

\myproc{\AddEdges{$\mathcal{C}$, $d$, $p$, $t$}}{
    $\mathcal{E}_E \gets \{\}$
    
    $\gamma \gets \{\}$
    
    \For{$\left\{ u, v \right\} $ \textbf{in} $\mathcal{C}$}{
        $i \gets \max\left\{  1,\frac{d_{u,v}}{p}\right\} $

	$n \gets$ \textbf{ShortestPath}($u$,$v$) \textbf{in} $\mathcal{E}_E$ \textbf{with edge weights} $\gamma$  \tcp*{infinity if no path exists}

	\If{$n > it$}{
            $\mathcal{E}_E \gets \mathcal{E}_E \cup \left\{\left\{ u, v \right\} \right\}$

	    $\gamma_{\left\{ u, v \right\} } \gets \lceil i \rceil$
        }
    }
    \KwRet $\mathcal{E}_E, \gamma$;
}
\myproc{\EdgeDepthFirstSearch{$\mathcal{E}_E$, $O$, $u$}}{
    \For{$v \in \left\{v \in \mathcal{V}_E | \left\{ u, v \right\}  \in \mathcal{E}_E \right\} $}{
        \If{$\left\{ u, v \right\}  \notin O$}{
            $O \gets $ \textbf{append}(O, $\left\{ u, v \right\} $)
            
            $O \gets $ \textbf{concatenate}(O, \EdgeDepthFirstSearch{$\mathcal{E}_E$, $O$, $v$})
        
        }
    }
    \KwRet $O$;
}

\myproc{\RefineEdges{$O$, $d$, $\gamma$, $p$}}{

    $\eta \gets 0$;

    \For{$\left\{ u, v \right\} $ \textbf{in} $O$}{
        $i \gets \max\left\{  1,\frac{d_{u,v}}{p} \right\} $

        \eIf{$|\lceil i \rceil - i + \eta| < |\lfloor i \rfloor - i + \eta|$}{
            $\gamma_{\left\{ u, v \right\} } = \lceil i \rceil$
            
            $\eta \gets \lceil i \rceil - i + \eta$
        }{
            $\gamma_{\left\{ u, v \right\} } = \lfloor i \rfloor$
            
            $\eta \gets \lfloor i \rfloor - i + \eta$
        }
    }

    \KwRet $\gamma$ 
}





\caption{Adding edges to elastic graphs}
\label{alg:add_edges}
\end{algorithm}

\subsection{Efficient inference with the object-specific MAM}

In our formulation, we use $\mathcal{M}$ to model the presence of a particular training letter at a certain location on the image. In practice, only a small number of training letters at a small number of locations would fit a given image well. To make inference efficient, we use fast feedforward operations to identify the rough locations of a small number of training letters (in the form of a small number of elastic graphs with different anchor points and small $\eta$), and do MPBP inference only with the relevant part and attention variables. All the hyper-parameters used in this section are empirically set with a validation set consisting of $1000$ \textit{easy} cABC images. The same set of hyper-parameters are then applied to all 3 difficulty levels (\textit{easy, medium} and \textit{hard}) without change.

We follow notations in Sec. 3.2 in the main text and in Sec.~\ref{sec:map_updates}. For a given $M\times M$ image, use $m_{I_{i j}}$ to denote the evidence coming from the image to the pixel variable $I_{i j}, 1 \le  i, j \le M$. For a given elastic graph $\mathcal{G}_E(\eta) = (\mathcal{V}_E, \mathcal{E}_E, \eta)$, we define its \textit{score} to be
\[
	\frac{1}{\left| \mathcal{V}_E \right|^{0.76}}\left(\sum_{v\in \mathcal{V}_E}\max_{\tilde{r}_v, \tilde{c}_v: ~ 0\le \tilde{r}_v - r_v, ~ \tilde{c}_v - c_v< \eta}\sum_{(i, j) \in I(x_{v \tilde{r}_v \tilde{c}_v}^{\mathcal{G}_E(\eta)})} m_{I_{i j}}\right)  
\]
Intuitively, to get the score of an elastic graph $\mathcal{G}_E(\eta)$, for each object part $v \in \mathcal{V}_E$, we obtain the score of the object part at a given location $(\tilde{r}_v, \tilde{c}_v)$ by summing up the evidence from pixels that are part of this object part, i.e. $\sum_{(i, j)\in I(x_{v \tilde{r}_v \tilde{c}_v}^{\mathcal{G}_E(\eta)})} m_{I_{i j}}$. We maximize over all the allowed locations $0 \le  \tilde{r}_v - r_v, \tilde{c}_v - c_v < \eta$ of the object part $v \in \mathcal{V}_E$ to get the score of the object part $\max_{\tilde{r}_v, \tilde{c}_v: 0\le \tilde{r}_v - r_v, \tilde{c}_v - c_v< \eta}\sum_{(i, j) \in I(x_{v \tilde{r}_v \tilde{c}_v}^{\mathcal{G}_E(\eta)})} m_{I_{i j}}$, and we sum over all the object parts in the elastic graph to get the unnormalized score of the elastic graph $\sum_{v\in \mathcal{V}_E}\max_{\tilde{r}_v, \tilde{c}_v: ~ 0\le \tilde{r}_v - r_v, ~ \tilde{c}_v - c_v< \eta}\sum_{(i, j) \in I(x_{v \tilde{r}_v \tilde{c}_v}^{\mathcal{G}_E(\eta)})} m_{I_{i j}} $. We normalize the score by $\frac{1}{\left| \mathcal{V}_E \right|^{0.76} }$ to prevent the process from being dominated by elastic graphs with a large number of vertices.

In our experiments, we first downsample the original $350\times 350$ images to $256\times 256$ images (i.e.\ $M = 256$), and use circles of radius $4$ as our only object part. We define $m_{i j} = 1.0$ for the pixel $(i, j)$ that is \texttt{ON}, and $m_{i j} = -0.89$ for the pixel $(i, j)$ that is \texttt{OFF}. To make the elastic graphs scores informative, instead of using $\eta = 256$ to allow the vertices to move around as much as possible, we use $\eta = 15$ to evaluate how well an elastic graph fits in a local region. To cover the entire image, we shift the anchor points from $(1, 1)$ to the right/downwards with a step size of  $13$, i.e. we consider the elastic graph  $ \mathcal{G}_E(15)$ and its transformed copies with anchor points $(1 + 13 \times k, 1 + 13 \times \ell), k, \ell \in \mathbb{N}$. This implies that, for each elastic graph $\mathcal{G}_E(15)$, we are evaluating its scores at 400 different locations.

We evaluate the scores for all the elastic graphs and their transformed copies. For a given anchor point, we rank the elastic graphs with this anchor point based on their scores, and keep the top $3$ elastic graph at each anchor point. We aggregate the top $3$ elastic graphs from all  $400$ anchor points, and keep the top $140$ (across all the anchor points) as estimates of the locations of the most promising elastic graphs.

We take the $140$ most promising elastic graphs and use $\eta=25$ to allow more invariance in the MAM inference. To parallelize the inference process, for each elastic graph $\mathcal{G}_E(25)$, we use $\sum_{(i, j) \in I(x_{v \tilde{r}_v \tilde{c}_v}^{\mathcal{G}_E(25)})} m_{I_{i j}}$ as the evidence for the part variable $x_{v \tilde{r}_v \tilde{c}_v}^{\mathcal{G}_E(25)}$, and do MPBP inference in parallel for all the elastic graphs. From the MPBP decodings, for each $v\in \mathcal{V}_E$, we can determine its most likely location, which we denote as $(\hat{r}_v, \hat{c}_v)$. We define $I(\mathcal{G}_E(25)) = \cup_{v \in \mathcal{V}_E} I(x_{v \hat{r}_v \hat{c}_v}^{\mathcal{G}_E(25)})$ as the set of pixels associated with the elastic graph MPBP decoding. For each pixel $(i, j)\in I(\mathcal{G}_E(25))$, use $\mathcal{N}((i, j), \mathcal{G}_E(25)) = \sum_{v \in \mathcal{V}_E}\mathbbm{1}_{(i, j) \in I(x_{v \hat{r}_v, \hat{c}_v}^{\mathcal{G}_E(25)})}$ to denote the number of object parts in the MPBP decoding that covers the pixel $(i, j)$.

As a final inference step, we approximate MPBP inference with the sparsifier with a heuristic pairwise reasoning step. More concretely, we consider all ${140 \choose 2} = 9730$ possible pairs of elastic graphs. We define the score of a given pair of elastic graphs $\mathcal{G}_E^{(1)}(25)$ and $\mathcal{G}_E^{(2)}(25)$ as
\[
	\sum_{(i, j)\in I(\mathcal{G}_E^{(1)}(25)) \cup I(\mathcal{G}_E^{(2)}(25))} \left[ m_{I_{i j}} - \left( \mathcal{N}((i, j), \mathcal{G}_E^{(1)}(25)) + \mathcal{N}((i, j), \mathcal{G}_E^{(2)}(25)) - 1 \right) * 0.33  \right]
\] 
i.e. we sum up the evidence from all the involved pixels, while applying a small penalty for object parts that overlap with each other.

We pick the pair of elastic graphs with the highest score as our interpretation of the test image, and heuristically assign the two markers to classifier cABC test images.

\begin{figure}[h!]
    \centering
    \includegraphics[width=\textwidth]{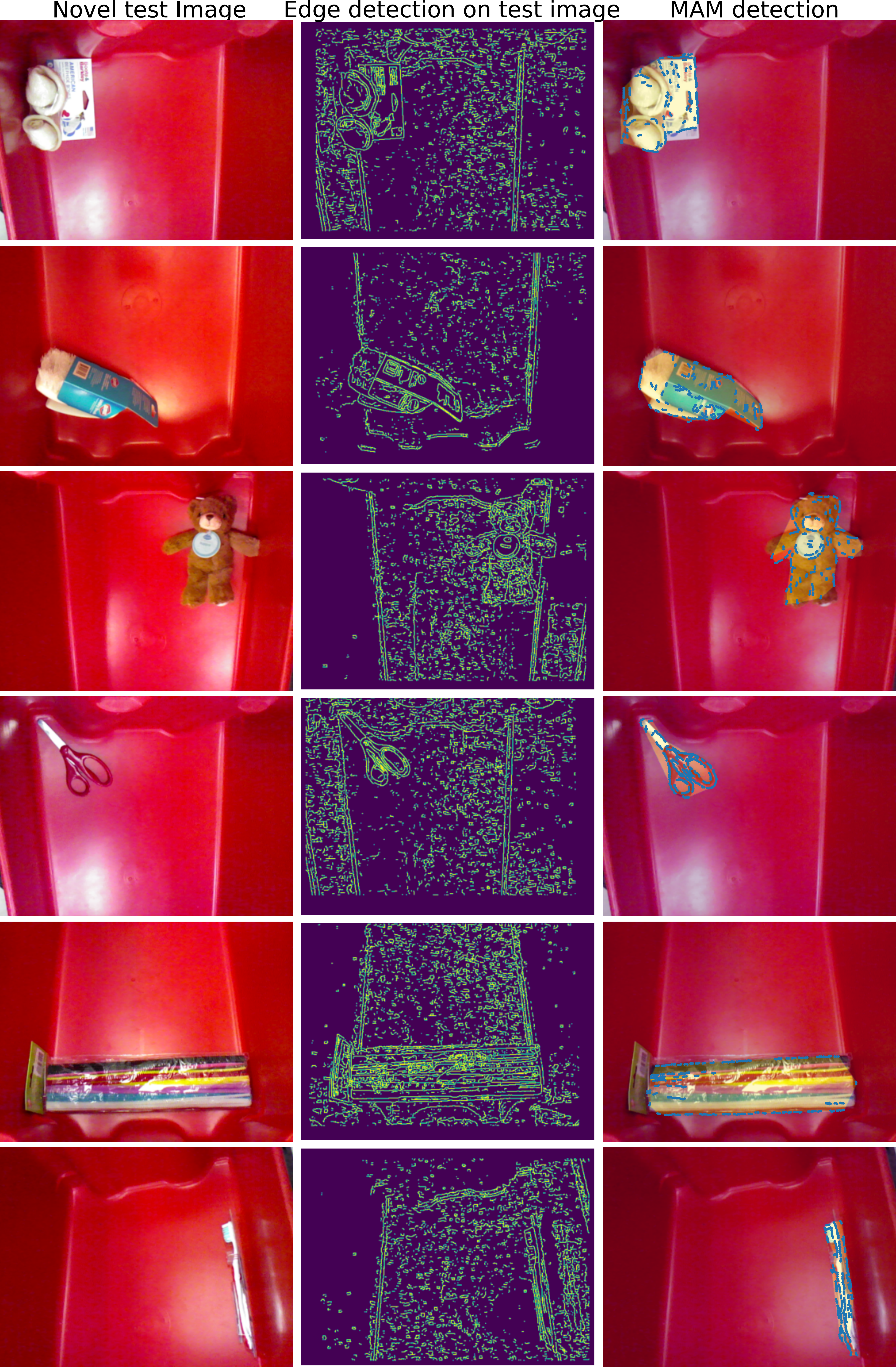}
    \caption{Visualization of MAM detections on realistic images for 6 objects (\emph{barkely\_hide\_bones, clorox\_utility\_brush, cloud\_b\_plush\_bear, fiskars\_scissors\_red, creativity\_chenille\_stems, oral\_b\_toothbrush\_red}) from the Object Segmentation Training Dataset. Blue dots in the right column represent object part activations from MPBP inference with MAMs.}
    \label{fig:apc}
    \vspace{-30pt}
\end{figure}

\section{Details for experiments on realistic images}
\subsection{Dataset}

We use realistic RGB-D images containing 39 different single objects in totes from the Object Segmentation Training Dataset~\cite{zeng2016multi}, available for download at \url{https://vision.princeton.edu/projects/2016/apc/}. For each object, we have multiple scenes. Each scene has the object in a different pose. For each scene, we have multiple frames. Each frame has the camera in a different pose. We use the RGB, depth and object masks associated with each frame as our train and test data. We randomly sample 70\% of the scenes as training data, and use the remaining 30\% as test data. Since we want to test an object-specific MAM modeling entire objects, we filter out frames in which the objects are only partially visible due to object and camera poses, by keeping frames with the largest mask sizes for each scene. For each training scene, we keep the top 3 frames. For each test scene, we keep the top 5 frames. This results in a dataset with 7797 training frames and 5870 test frames across all objects. See the left column in \textbf{Fig.~\ref{fig:apc}} for the visualization of some example images.

\subsection{Edge detection and heuristic sparsification}

For edge detection, we take the maximum response from grayscale filters of 16 orientations over the 3 channels of RGB images, followed by a local suppression step similar to that used in Canny edge detection. See the middle column in \textbf{Fig.~\ref{fig:apc}} for the visualization of the maximum response on test images from the detection of edges of 16 orientations.

For sparsification, we adopt a heuristic procedure, where we pick points roughly at equal distances. More concretely, we start from a random point, and greedily pick points at a distance close to our target distance, until we can no longer pick new points. From the set of unpicked points, we then randomly pick a new starting point that is at least the target distance away from already picked points, and continue the greedy procedure. We repeat until we can no longer find new points.

\subsection{Visualization of results}

See the right column in \textbf{Fig.~\ref{fig:apc}} for the visualization of MAM detections on 6 objects from the Object Segmentation Training Dataset. Blue dots represent object part activations from MPBP inference with MAMs. From the results we can see that, dispite challenges like deformable objects, similar color to background, and adversial surface properties (e.g. reflective), our object-specific MAMs do a good job at detecting the objects, and give interpretable decodings, which we can use to further understand the pose of the detected objects.

\bibliographystyle{plain}
\bibliography{refs}